\pdfoutput=1

\documentclass{article} 
\usepackage{collas2024_conference,times}
\usepackage{easyReview}
\collasfinalcopy

\usepackage{amsmath,amsfonts,bm}

\def\eqref#1{equation~\ref{#1}}

\def\1{\bm{1}}

\DeclareMathAlphabet{\mathsfit}{\encodingdefault}{\sfdefault}{m}{sl}
\SetMathAlphabet{\mathsfit}{bold}{\encodingdefault}{\sfdefault}{bx}{n}

\usepackage{hyperref}
\hypersetup{
    colorlinks=true,
    linkcolor=red,
    filecolor=magenta,
    urlcolor=blue,
    citecolor=purple,
    pdftitle={Overleaf Example},
    pdfpagemode=FullScreen,
    }

\usepackage{times}  
\usepackage{helvet}  
\usepackage{courier}  
\usepackage{graphicx} 
\usepackage{caption} 
\usepackage{subfigure}

\usepackage{times}
\usepackage{graphicx}
\usepackage{amsmath}
\usepackage{amssymb}
\usepackage{wrapfig}
\usepackage{bm}
\usepackage{multirow}
\usepackage{enumitem}

\usepackage{algorithm}
\usepackage{algorithmic}

\usepackage{newfloat}
\usepackage{listings}

\title{Towards Diverse Evaluation of Class Incremental Learning: A Representation Learning Perspective}

\author{Sungmin Cha$^{1}$, Jihwan Kwak$^{2}$, Dongsub Shim$^{3}$, Hyunwoo Kim$^{4}$, \\ \textbf{Moontae Lee}$^{3,5}$, \textbf{Honglak Lee}$^{3}$, \textbf{and} \textbf{Taesup Moon}$^{2,6}$\thanks{Corresponding author} \\
$^1$Computer Science Department at the Courant Institute of Mathematical Sciences, New York University\\
$^2$Department of Electrical and Computer Engineering, Seoul National University\\
$^3$Advanced Machine Learning Lab, LG AI Research\\
$^4$Zhejiang Lab\\
$^5$Department of Information and Decision Sciences, University of Illinois Chicago\\
$^6$ASRI / INMC / IPAI / AIIS, Seoul National University\\
\texttt{sungmin.cha@nyu.edu}, \
\texttt{\{kkwakzi, tsmoon\}@snu.ac.kr}, \
\texttt{hwkim@zhejianglab.com},  \\ \texttt{\{dongsub.shim, moontae.lee, honglak.lee\}@lgresearch.ai}
}

\begin{document}

\maketitle

\begin{abstract}

Class incremental learning (CIL) algorithms aim to continually learn new object classes from incrementally arriving data while not forgetting past learned classes. The common evaluation protocol for CIL algorithms is to measure the average test accuracy across all classes learned so far --- however, we argue that solely focusing on maximizing the test accuracy may not necessarily lead to developing a CIL algorithm that also continually learns and updates the representations, which may be transferred to the downstream tasks.  To that end, we experimentally analyze neural network models trained by CIL algorithms using various evaluation protocols in representation learning and propose new analysis methods. Our experiments show that most state-of-the-art algorithms prioritize high stability and do not significantly change the learned representation, and sometimes even learn a representation of lower quality than a naive baseline. However, we observe that these algorithms can still achieve high test accuracy because they enable a model to learn a classifier that closely resembles an estimated linear classifier trained for linear probing. Furthermore, the base model learned in the first task, which involves single-task learning, exhibits varying levels of representation quality across different algorithms, and this variance impacts the final performance of CIL algorithms. Therefore, we suggest that the representation-level evaluation should be considered as an additional recipe for more diverse evaluation for CIL algorithms.

\end{abstract}

\section{Introduction}

Neural networks have achieved great success in various fields such as computer vision, natural language processing, and reinforcement learning~\citep{(deeplearning1)lecun2015deep, (deeplearning2)bengio2021deep}. Among them, image classification is the first representative task leading to the significant progress of neural networks~\citep{(imagenet)deng2009imagenet, (cifar)krizhevsky2009learning, (adam)kingma2014adam}. Furthermore, the ImageNet~\citep{(imagenet)deng2009imagenet} pretrained model has been widely used as an initial model for transfer learning in other downstream tasks, such as object detection, semantic segmentation, and other classification datasets. In terms of representation learning, experimental analyses have shown that models achieving better classification accuracy learn better quality of representations, leading to better ability for transfer learning to various downstream tasks~\citep{(transferlearning)kornblith2019better}.

\begin{figure}[t]
\centering 
{\includegraphics[width=0.85\linewidth]{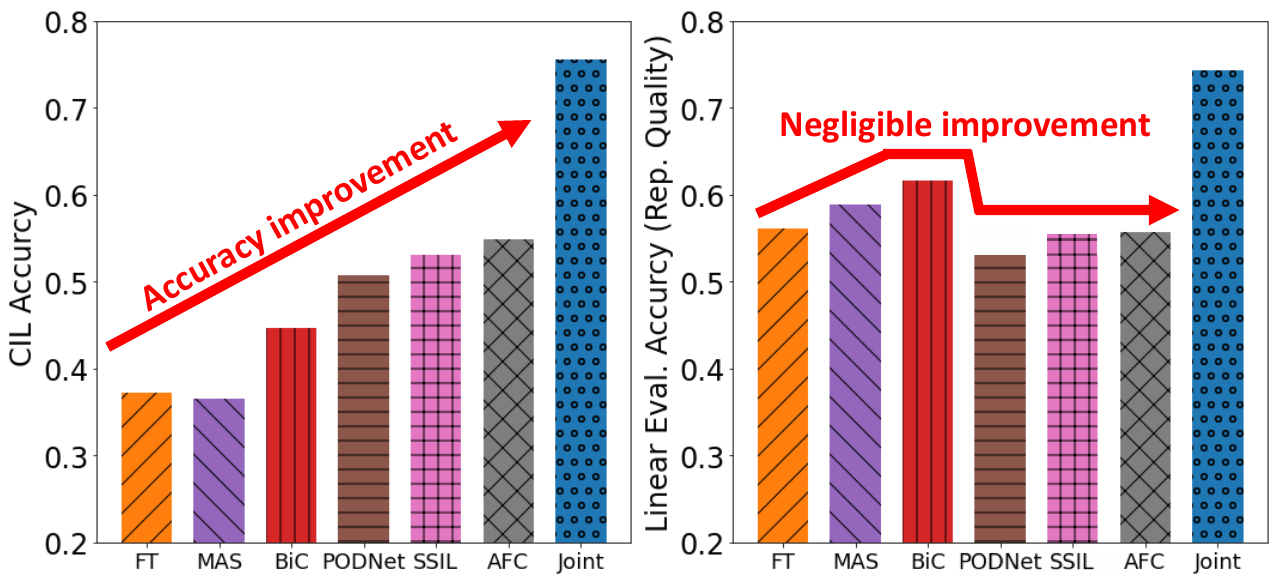}
\label{figure:sup_base_evcil}}
\caption{{Experimental results of CIL using the ImageNet-100 dataset for a 10-tasks scenario. The accuracy of state-of-the-art {regularization-based} CIL algorithms have been gradually increasing, approaching that of Joint training (left). However, we experimentally confirm that the improvement of the quality of representations learned by them is negligible or even worse than naive baselines (right).}}
\vspace{-.1in}
\label{figure:intro_figure}
\end{figure}


However, neural networks exhibit a substantial gap with humans in their ability to continually learn from a series of tasks. 
To narrow this gap, research on continual learning (CL) has started, starting with image classification as the primary task~\citep{(cl_survey1)parisi2019continual, (cl_survey2)delange2021continual, (cil_survey)masana2020class}. 
Among the three scenarios of CL, Class Incremental Learning (CIL) is considered as an important sub-category that has garnered significant attention~\citep{(CL_scenarios)van2019three, (cil_survey)masana2020class}. This scenario models a practical scenario that can be encountered in many real-world applications and is considered as the hardest compared to the other scenarios~\citep{(CL_scenarios)van2019three}, as the task-id is not available at inference time. 
In this CIL, the goal of a learning agent is  to successfully integrate knowledge gained from new object classes (plasticity) from incrementally arriving data while overcoming catastrophic forgetting for knowledge of the past learned classes (stability). However, achieving this goal is difficult  due to neural networks suffer from a trade-off between stability and plasticity~\citep{(tradeoff)mermillod2013stability}.

The effectiveness of CIL algorithms is evaluated based on the average test accuracy across all the classes learned so far since it is regarded as a good proxy for measuring both plasticity (for learning new classes) and stability (for not forgetting past classes). Recently proposed CIL algorithms have aimed to increase the average test accuracy after learning the final task~\citep{(cil_survey)masana2020class}. As presented in Figure \ref{figure:intro_figure} (left), the regularization-based methods using the exemplar memory have achieved the sound progress in terms of test accuracy improvement, even approaching the performance of the model jointly trained with the entire training dataset~\citep{(bic)wu2019large, (ss-il)ahn2021ss, (lucir)hou2019learning, (podnet)douillard2020podnet, (afc)kang2022class}. 
Similar to single-task training (\textit{e.g.}, ImageNet training), high test accuracy of a trained model is regarded as an indicator of a better model in CIL. However, the evaluation of the representations learned by state-of-the-art CIL algorithms has not been widely discussed so far. Therefore, it remains unclear whether their performance gain comes from continually learning better representations or from other factors.


In this paper, we argue that the horse race toward simply maximizing the average test accuracy has limitations and may not necessarily lead to the development of effective CIL algorithms. Therefore, we raise the necessity to evaluate the quality of learned representations to diversify the evaluation of CIL. Our motivation comes from the experimentally confirmed relationship between the quality of representations and the classification accuracy of the classification model~\citep{(transferlearning)kornblith2019better}. 
Unlike \cite{davari2022probing}, which analyzed forgetting of representations in continual learning using a naive baseline such as finetuning, we evaluate and analyze the representations learned by state-of-the-art regularization-based CIL algorithms.
To evaluate learned representations by CIL algorithms, first, we borrow two evaluation protocols of representation learning to solely evaluate the quality of encoders learned by the CIL algorithms: 1) fix the encoder and re-train the final linear layer or run the $k$-nearest neighbor (NN) classifier using the entire training set and check the test accuracy, and 2) perform transfer learning with the incrementally learned encoder to downstream tasks and report the test accuracy on those tasks.
Second, to check the level of changes of the representations, we report the level of changes of representations via CKA (Centered Kernel Alignment) measure~\citep{(cka)kornblith2019similarity}. 
Additionally, we devise a metric to evaluate how closely the learned output layer weights of each CIL algorithm resemble those of a linear classifier trained for linear probing.
By testing with above evaluation and analysis protocol on class incrementally learning ImageNet-100 in two major CIL scenarios (\textit{e.g.}, 10 and 11 tasks), we obtain the following findings: 

\begin{itemize}
  \item {First, {despite achieving high test accuracy, state-of-the-art regularization-based CIL algorithms end up learning representations that are either inferior or comparable to those of other baselines that achieve lower test accuracy.}}
  \item {Second, {the majority of state-of-the-art regularization-based algorithms prioritize stability, leading to minimal enhancement in representation during CIL. Additionally, we confirm that their superior performance may stem from learning an output layer that closely resembles that of a linear classifier trained using linear probing.}}
  \item {Third, the representation quality of the first task model, which is single-task learning and not heavily influenced by CIL algorithms, can vary among algorithms and significantly impact their final performance.}
\end{itemize}


\section{Related Work}

\noindent\textbf{Supervised class-incremental learning} \ \
Continual learning (CL) methods can be classified into three types~\citep{(cl_survey2)delange2021continual}: dynamic architecture-based approaches, regularization-based methods, and exemplar-based methods. Dynamic architecture-based approaches extend the capacity of neural networks dynamically to learn a new task without catastrophic forgetting~\citep{(PNN)RusuRabiDesjSoyeKirk2016, (imagenet)deng2009imagenet, (Packnet)MallyaLazebnik18, (ProgressCompress)SchwarzLuketinaHadsell18, hung2019compacting, (CN-DPM)lee2020neural}. Regularization-based methods maintain important weights for previous tasks during training to prevent catastrophic forgetting, showing superior performance, especially for task-incremental learning~\citep{(ewc)kirkpatrick2017overcoming, (MAS)aljundi2018memory, (Rwalk)chaudhry2018riemannian, (UCL)ahn2019uncertainty, (AGS-CL)jung2020adaptive, (widelocal_cl)mirzadeh2020understanding, (cpr)cha2021cpr}. However, they exhibit degraded performance for class-incremental learning (CIL)~\citep{(CL_scenarios)van2019three}. Exemplar-based methods store a subset of previous task data as exemplars and retrieve them when training a new task, showing superior performance in most CIL scenarios~\citep{(icarl)rebuffi2017icarl, (eeil)castro2018end, (bic)wu2019large, (gdumb)prabhu2020gdumb}. They have shown even better performance when combined with distillation-based methods~\citep{(podnet)douillard2020podnet, (lucir)hou2019learning, (ss-il)ahn2021ss, (afc)kang2022class, (prototype)asadi2023prototype} {as well as a contrastive learning-based method~\citep{cha2021co2l}}. Additionally, some works pointed out the problem of normalization layers in CIL and proposed a novel normalization layer designed for CIL~\citep{(cn)pham2022continual, (tbbn)cha2023rebalancing}.
Recently, CIL with pretrained models has gained attention, leveraging the superior representations of these models for CIL without relying on exemplar memory~\citep{panos2023first, zhang2023slca, (l2p)wang2022learning}.


\noindent\textbf{Analysis of learned representations by CIL} \ \
Studies probing the quality of learned representations in CIL have yielded insightful observations. \cite{yu2020semantic} empirically demonstrates that CIL with the encoder alone (using metric learning-based finetuning) exhibited limited forgetting, while combining the encoder and output layer (via cross-entropy-based finetuning) led to more substantial forgetting across the network. 
\cite{vogelstein2020representation} highlights the limitation associated with accuracy-based evaluation in  lifelong learning and introduces several metrics that provide a more comprehensive assessment.
In the realm of task-incremental learning (TIL), \cite{davari2022probing} embarks on an in-depth exploration of representation quality. Notably, they confirm that representation forgetting under naive finetuning is less pronounced than the corresponding accuracy drop in TIL. Furthermore, they demonstrate that contrastive learning-based loss functions exhibit enhanced resilience against representation forgetting, aligning with insights from continual self-supervised learning~\citep{(cassle)fini2022self, (continuity)madaan2022representational}.

In contrast to previous studies, our study distinguishes itself in two key aspects. First, we perform extensive experiments to assess the learned representations using state-of-the-art {regularization-based} CIL algorithms. Second, drawing from our experimental results, we highlight a common drawback in current CIL research, which tends to concentrate solely on maximizing classification accuracy. In light of these findings, we advocate for diversified evaluation methods to more effectively evaluate the quality of representations learned by CIL algorithms.

\section{{Towards Diverse Evaluation of CIL from a Representation Perspective}}

\begin{figure*}[t]
\centering 
\includegraphics[width=0.98\linewidth]{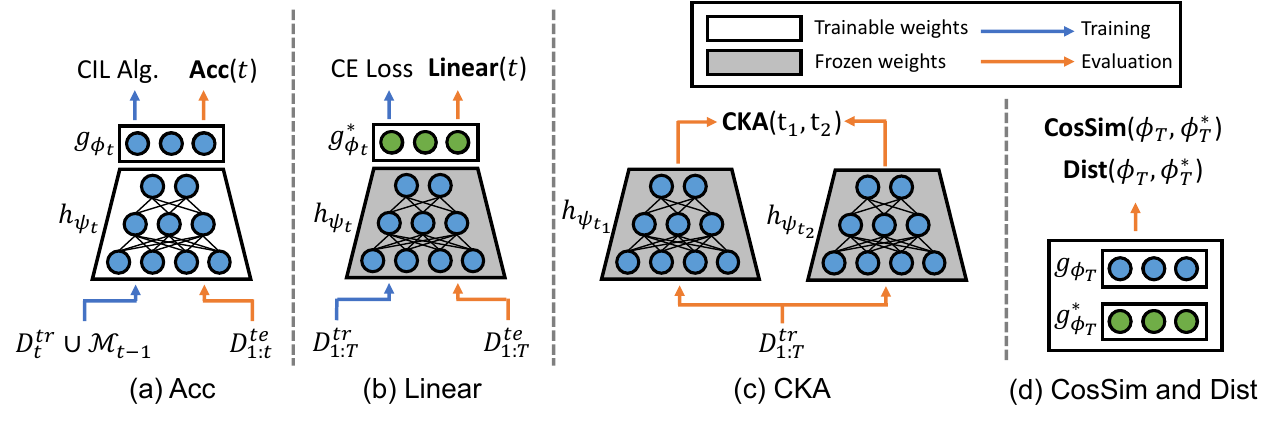}
\vspace{-.1in}
\caption{{This figure illustrates proposed and used evaluation methods in this paper. (a): the standard evaluation method that measures classification accuracy on test data following training task $t$. (b): linear probing evaluation for a model trained on task $t$. (c): measuring CKA between representations of two models trained on different tasks. (d): a comparison of weights in the output layer.}}
\vspace{-.1in}
\label{figure:method}
\end{figure*}

\subsection{Problem formulation and preliminaries}\label{sec:three_scenarios}

In this section, we {briefly introduce the preliminaries and problem formulation of our paper.} {We} follow the {general} settings and problem formulation of class-incremental learning (CIL) proposed in previous papers~\citep{(icarl)rebuffi2017icarl, (CL_scenarios)van2019three,(cil_survey)masana2020class}.

\noindent\textbf{Notations and settings} \ \
We assume a sequential task setting, where $ t \in \{1, \dots, T\}$ represents the $t\textsuperscript{th}$ task. Task-specific training and test datasets {at task $t$} are denoted as $D_{t}^{\mathrm{tr}}$ and $D_{t}^{\mathrm{te}}$, respectively. Each task-specific dataset $D_t=\{(\bm x^{(i)}_t, y^{(i)}_t)\}^{N}_{i=1}$ consists of $N$ pairs of an input image and its target label. The target label $y_t$ is assumed to be sampled from a task-specific class set $\mathcal{C}_{t}$ {which are disjoint across different tasks, \textit{i.e.} $\mathcal{C}_{i}\cap\mathcal{C}_{j}=\emptyset \text{ }\forall i\neq j  \in \{1, \dots, T\}$. Exemplar memory is allocated to store and replay a small number of data instances of previous tasks. More specifically, exemplar memory which holds data seen until task $t-1$ is denoted as $\mathcal{M}_{t-1}$ and is used for training at task $t$. In this paper, we consider a class-balanced memory which is simple in that it stores equal number of images per class and is known to be efficient~\citep{(gdumb)prabhu2020gdumb, (eeil)castro2018end, (lucir)hou2019learning, (podnet)douillard2020podnet}.} 

{At task $t$, a classification model  $f_{\bm \theta_{t}}= (g_{\bm \psi_{t}} \circ h_{\bm \phi_{t}})$ is trained, where $h_{\bm \psi_{t}}$ and $g_{\bm \phi_{t}}$ indicates the encoder and the output layer of the model, respectively. In this paper, we consider and compare CIL algorithms that use the cross-entropy (CE) objective function as a main training objective. Note that the model trained with the entire training datasets until task $t$ is denoted as \textit{joint} which is an oracle case.}
{$f_{\bm \theta_{t}}$ is trained on $D_{t}^{\mathrm{tr}} \cup \mathcal{M}_{t-1}$ for multiple epochs (offline training) at task $t$, and evaluated on $D_{1:t}^{\mathrm{te}}$. In CIL, task ID, an additional supervisory signal, is not provided as it adopts a shared output layer.} 

\noindent{\textbf{{Conventional} metrics of CIL} \ \
Among various ordinary metrics for CIL, 
we adopt two general metrics for evaluating the performance of CIL algorithm~\citep{(cil_survey)masana2020class}: \textbf{Acc}${(t)}$ (shown in Figure \ref{figure:method}) and \textbf{AvgAcc}${(t)}$. \textbf{Acc}${(t)}$ is the test accuracy of $f_{\bm \theta_{t}}$ on $D_{1:t}^{\mathrm{te}}$, and \textbf{AvgAcc}${(t)}$ is the average of \textbf{Acc}${(t)}$ from the first task to the $t$-th task, {\textit{i.e.,} $\text{\textbf{AvgAcc}}{(t)}=\frac{1}{t}\sum_{l=1}^{t}$\textbf{Acc}${(l)}$.}}

\subsection{{Proposed Evaluation method for analysis from a representation perspective}}\label{sec:evaluation protocol}

To comprehensively assess the improvements in learned representations by CIL algorithms, we deploy a structured evaluation framework encompassing both in-domain and out-domain perspectives.

\noindent{{(1)\ \textbf{In-domain evaluation: \textbf{Linear}${(t)}$ and \textbf{$k$-NN}${(t)}$}}} \ \
To compare the improvement of representations learned by CIL algorithms, {we borrow the evaluation methods used in representation learning research ~\citep{(barlow_twins)zbontar2021barlow, (moco)he2020momentum}: Linear evaluation  and $k$-NN classification. As shown in Figure \ref{figure:method} (b), representations of each encoder $h_{\bm \psi_{t}}$ is evaluated by freezing the encoder $h_{\bm \psi_{t}}$ and conducting a linear probing by re-training the final linear layer to obtain an estimated linear classifier $g^{*}_{\bm \phi}$. $k$-NN classifier ($k=20$) is also constructed with the frozen encoder}. Note that the entire training dataset of a given CIL scenario, $D_{1:T}^{\mathrm{tr}}$, is used to train the estimated linear classifier or to formulate $k$-NN classifier and that the is entire test dataset $D_{1:T}^{\mathrm{te}}$ is used for evaluation.

\noindent{{(2)\ \textbf{Out-domain evaluation: \textbf{CLS}${(t)}$}}} \ \
To further evaluate the quality of the learned representations in more general aspects, we conduct experiments of transfer learning with out-domain datasets as well. We consider three downstream tasks of classification, namely STL-10~\citep{(stl10)coates2011analysis}, CUB200~\citep{(cub200)wah2011caltech}, and resized ($96\times96$) CIFAR-10~\citep{(cifar)krizhevsky2009learning}. For each encoder $h_{\bm \psi_{t}}$, we perform linear evaluation using each dataset and report their average classification accuracy.

\noindent{(3)\ {\textbf{Representation similarity comparison: \textbf{CKA}${(t_{1},t_{2})}$}}} \ \
We compare the degree of changes of learned representations during a task change in CIL from $t_{1}$ to $t_{2}$ by measuring their similarity using CKA~\citep{(cka)kornblith2019similarity}. That is, as shown in Figure \ref{figure:method} (c), we measure CKA between $h_{\bm \psi_{t_{1}}}$ and $h_{\bm \psi_{t_{2}}}$ by using entire training dataset $D_{1:T}^{\mathrm{tr}}$.

\noindent{{(4)\ \textbf{{Comparison of weights of output layer}}}} \ \ 
{For further analysis for output layer, we propose to conduct comparison between a classifier layer $g_{\bm \phi_T}$ trained during CIL and an estimated linear classifier $g^{*}_{\bm \phi_T}$ trained for linear probing~\citep{davari2022probing}, for the final $T$-th task's model.}
Let $W\in\mathbb{R}^{D \times C}$ and $W^{*}\in\mathbb{R}^{D \times C}$ are parameters of the classifier layer of $g_{\bm \phi_T}$ and $g^{*}_{\bm \phi_T}$, respectively.
$D$ denotes the dimension of an output feature of $h_{\bm \psi_{T}}$ and $C$ stands for the number of the whole classes.
To compare both parameters, we calculate both cosine similarity and $L_2$ distance between them as below:
\begin{align*}
\mathrm{\textbf{CosSim}}(\phi_{T}, \phi^{*}_{T}) &= \frac{1}{C}\sum_{i}^{C} \frac{(W_i)^{\intercal}W^{*}_{i}}{\|W_i\|_{2}\|W^{*}_{i}\|_{2}} &
\mathrm{\textbf{Dist}}(\phi_{T}, \phi^{*}_{T}) &= \frac{1}{C}\sum_{i}^{C} \|W_i - W^{*}_{i}\|_{2}
\end{align*}
where $i$ denotes an index of column axis and $W_i, W^{*}_{i} \in \mathbb{R}^{D}$.
{Note that, when \textbf{CosSim}$(\phi_{T}, \phi^{*}_{T})$ is high and \textbf{Dist}$(\phi_{T}, \phi^{*}_{T})$ is low at the same time, it means the classifier layer is close to the estimated linear classifier.}


\section{Experimental Setup}

\noindent{\textbf{Baselines}} \ \ 
{As discussed in the Related Work section, class incremental learning (CIL) research has evolved in various forms, such as regularization-based, exemplar-based, and model expansion-based methods. Recently, approaches using contrastive learning and pretrained models have also been proposed. However, in this paper, we focus on analyzing and evaluating regularization-based methods that leverage exemplar memory. The rationale for this focus is twofold: 1) scontrastive learning-based method have already demonstrated gradual improvements in representation~\citep{cha2021co2l, (cassle)fini2022self} 2) CIL with a pretrained model starts with superior representations, leading to either freezing these models or aiming for minimal changes, which sets them apart from previous studies~\citep{panos2023first, zhang2023slca,(l2p)wang2022learning}. On the other hand, while regularization-based methods using the exemplar memory have been studied for a long time and continue to report excellent results, the analysis and experimentation regarding the learned representations of these methods are generally overlooked.}
The baseline algorithms used in our experiments and brief descriptions of them are as follows:

\begin{enumerate}[label=\arabic*)]
\item{Finetuning (FT): {It is a naive approach using fine-tuning a model with exemplars.}}
\item{MAS~\citep{(MAS)aljundi2018memory}: It measures the importance of each weight that constitutes the model using gradients and uses this importance as the strength of regularization to overcome catastrophic forgetting.}
\item{BiC~\citep{(bic)wu2019large}: It overcomes catastrophic forgetting by performing knowledge distillation from the previously learned model, as in LWF~\citep{(lwf)li2017learning}. Additionally, biased prediction issues in the output layer are resolved through post-processing on the prediction score. We also report results of BiC (w/o BC) which indicates results without the bias correction post-processing for output logits.} 
\item{PODNet~\citep{(podnet)douillard2020podnet}: It uses a more sophisticated spatial-based distillation loss to balance between learning new classes and forgetting previously learned classes.}
\item{SSIL~\citep{(ss-il)ahn2021ss}: It uses separated softmax and task-wise knowledge distillation to alleviate biased predictions.}
\item{AFC~\citep{(afc)kang2022class}: It calculates the importance in each feature map and proposes regularization using this importance to learn new knowledge well while preserving previously learned knowledge.}
\end{enumerate}
For our experiments, we train models using the official codes of PODNet, SSIL, and AFC for each algorithm. For BiC, we conduct experiments using the implementation in the official code of PODNet, and for FT and MAS, we conduct experiments using the CIL framework proposed in \citep{(cil_survey)masana2020class}.

\noindent{\textbf{CIL scenarios}} \ \
{We consider two CIL scenarios  using the ImageNet-100 dataset~\citep{(imagenet)deng2009imagenet}.} {The first scenario, denoted as \textbf{10-tasks}, is  consisting of 10 tasks each with 10 classes that are continuously learned. The second scenario, denoted as \textbf{11-tasks}, is a  scenario where 50 classes are learned in the base task (first task), followed by 10 continuous tasks each with 5 classes.}

\noindent{\textbf{Other  settings}} \ \
{
For all experiments, we use the ResNet-18 ~\citep{(resnet)he2016deep}.
All the baseline models are trained with the same hyperparameters proposed {by each algorithm}. 
For additional training {to get a result of \textbf{Linear}${(t)}$}, we train a new output layer with a mini-batch size of 256 with 30 epochs for in-domain dataset and 100 epochs for out-domain datasets. We use SGD optimizer with an initial learning rate of 0.1 and momentum of 0.9 and decay rate of 0.0001. We apply a schedule that multiplies the learning rate by 0.1 at \{10, 20\} and \{40, 80\} epochs for in-domain and out-domain evaluation, respectively.
We conduct experiments for three seeds and report averaged results of them.}

Note that, in the Appendix, we present the experimental setting and experimental results for alternative CIL algorithms, such as LUCIR, as well as for a distinct dataset, such as CIFAR-100.

\section{{Experimental Results with the proposed evaluation}}

{We then evaluate several regularization-based class incremental learning (CIL) algorithms using the proposed evaluation method. Our findings are summarized into three key points, and we present the experimental results sequentially.}

\subsection{{Achieving superior performance in conventional metrics does not always mean learning a superior representation}}

\begin{figure*}[t]
\centering 
\subfigure[{Acc}($t$).]
{\includegraphics[width=0.32\linewidth]{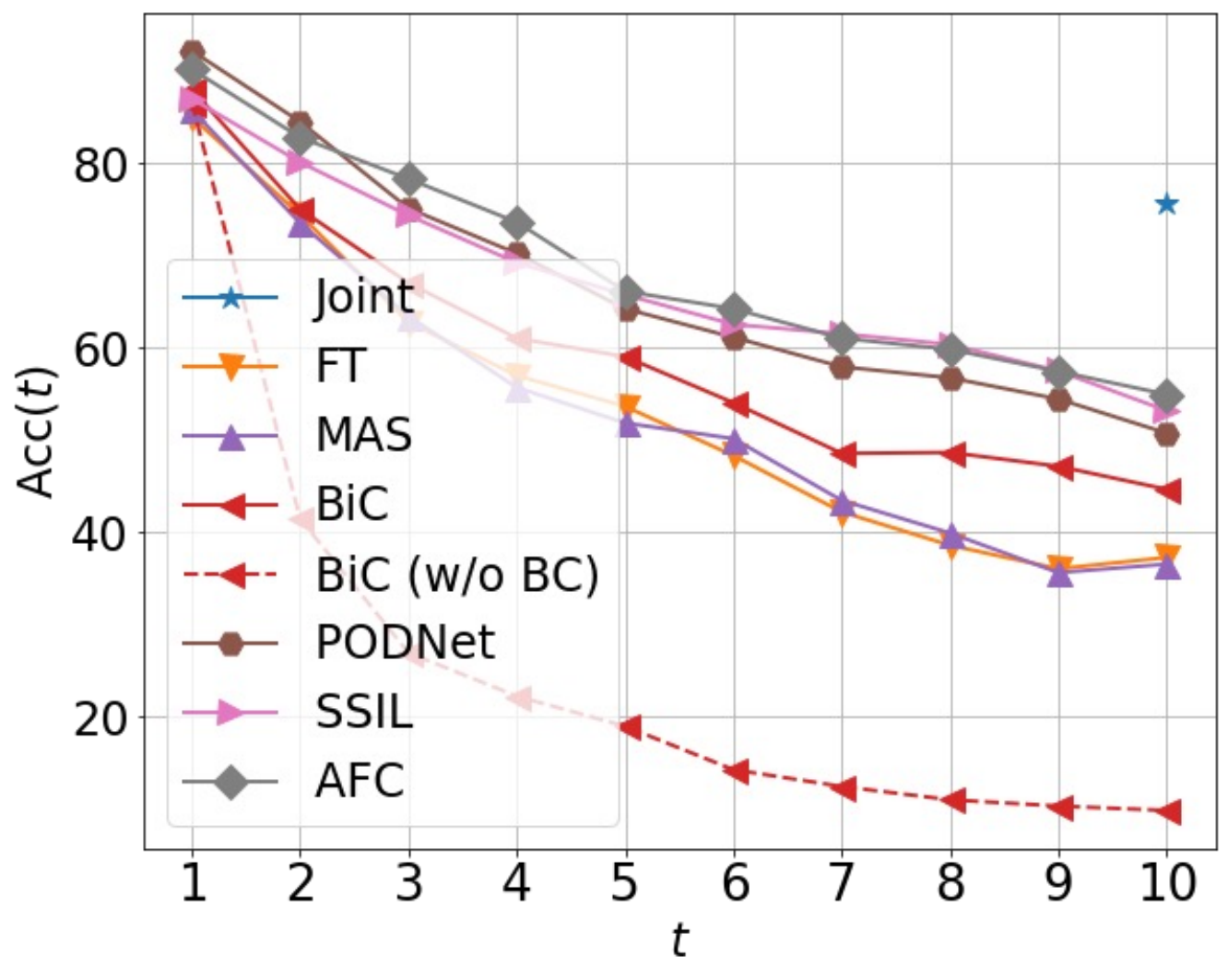}
\label{figure:10tasks_acc}}
\subfigure[$k$\textbf{-NN}($t$)]
{\includegraphics[width=0.32 \linewidth]{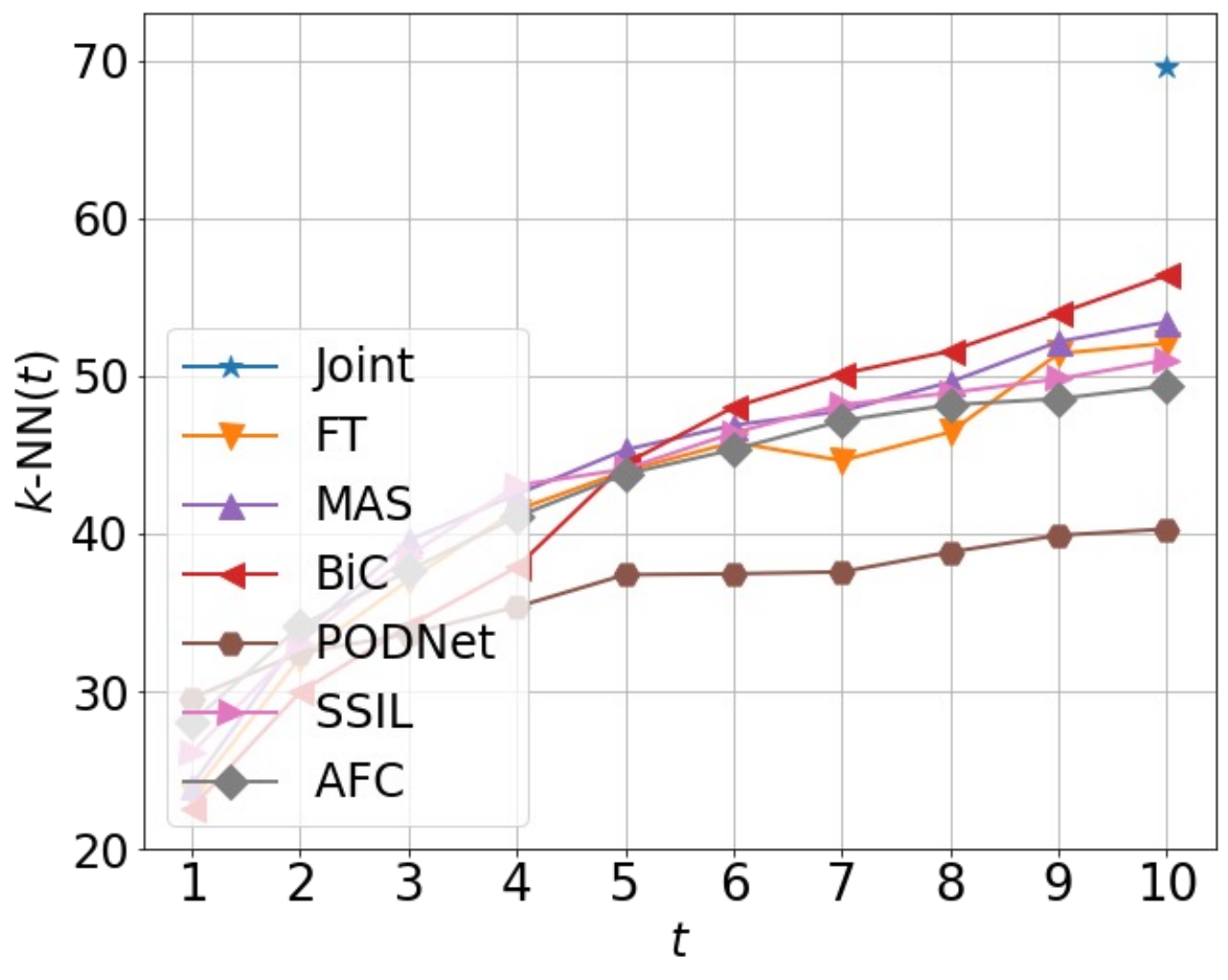}
\label{figure:10tasks_knn}}
\subfigure[All eval. metrics]
{\includegraphics[width=0.32 \linewidth]{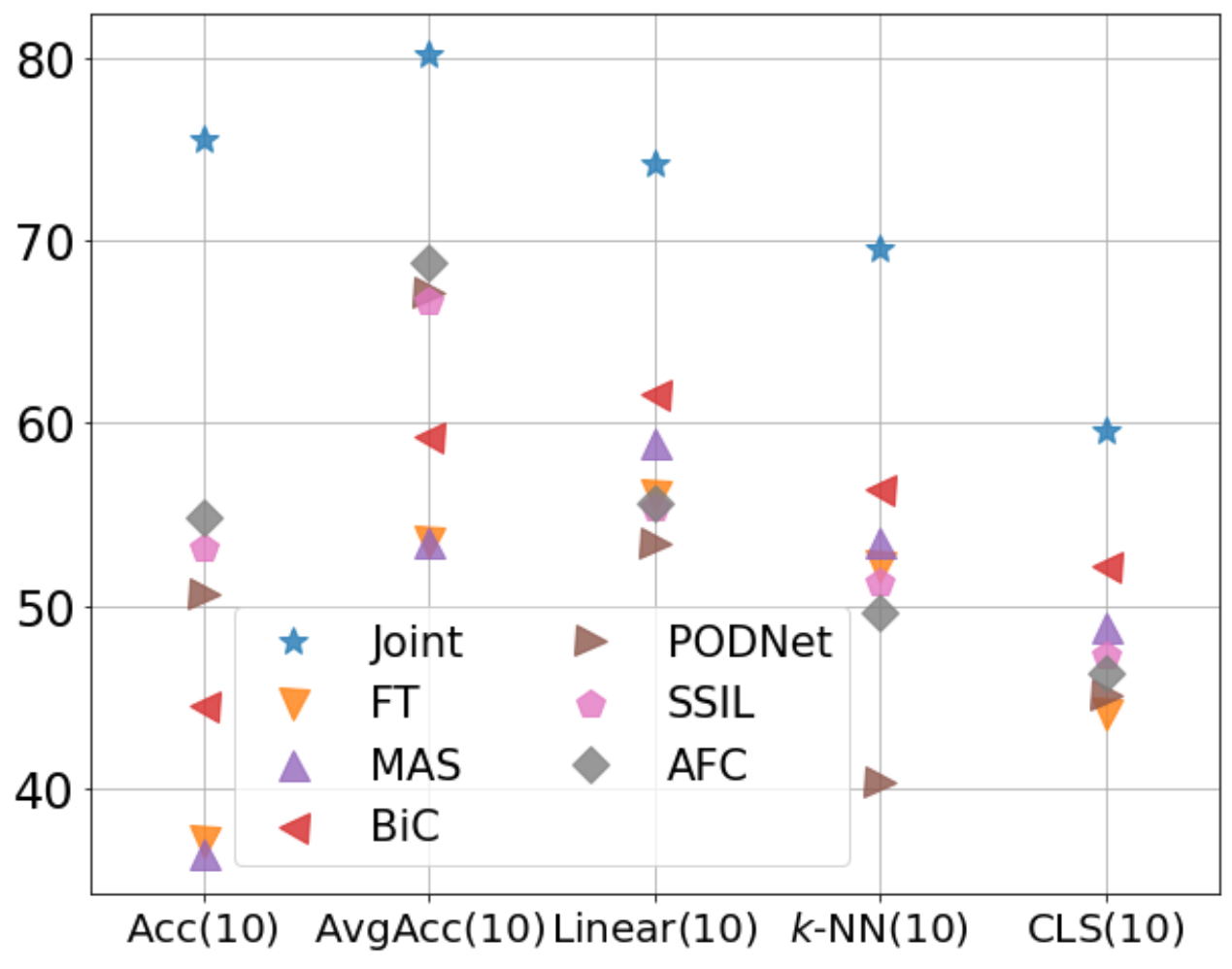}
\label{figure:10tasks_entire}}
\vspace{-.1in}
\caption{{The experimental results of regularization-based CIL algorithms for a 10-task scenario using the ImageNet-100 dataset. "Joint" refers to the performance of the upper bound case using the entire datasets.}}
\vspace{-.1in}
\label{figure:10tasks_figures}
 \end{figure*}

\begin{figure*}[!t]
\centering 
\subfigure[Acc($t$).]
{\includegraphics[width=0.32\linewidth]{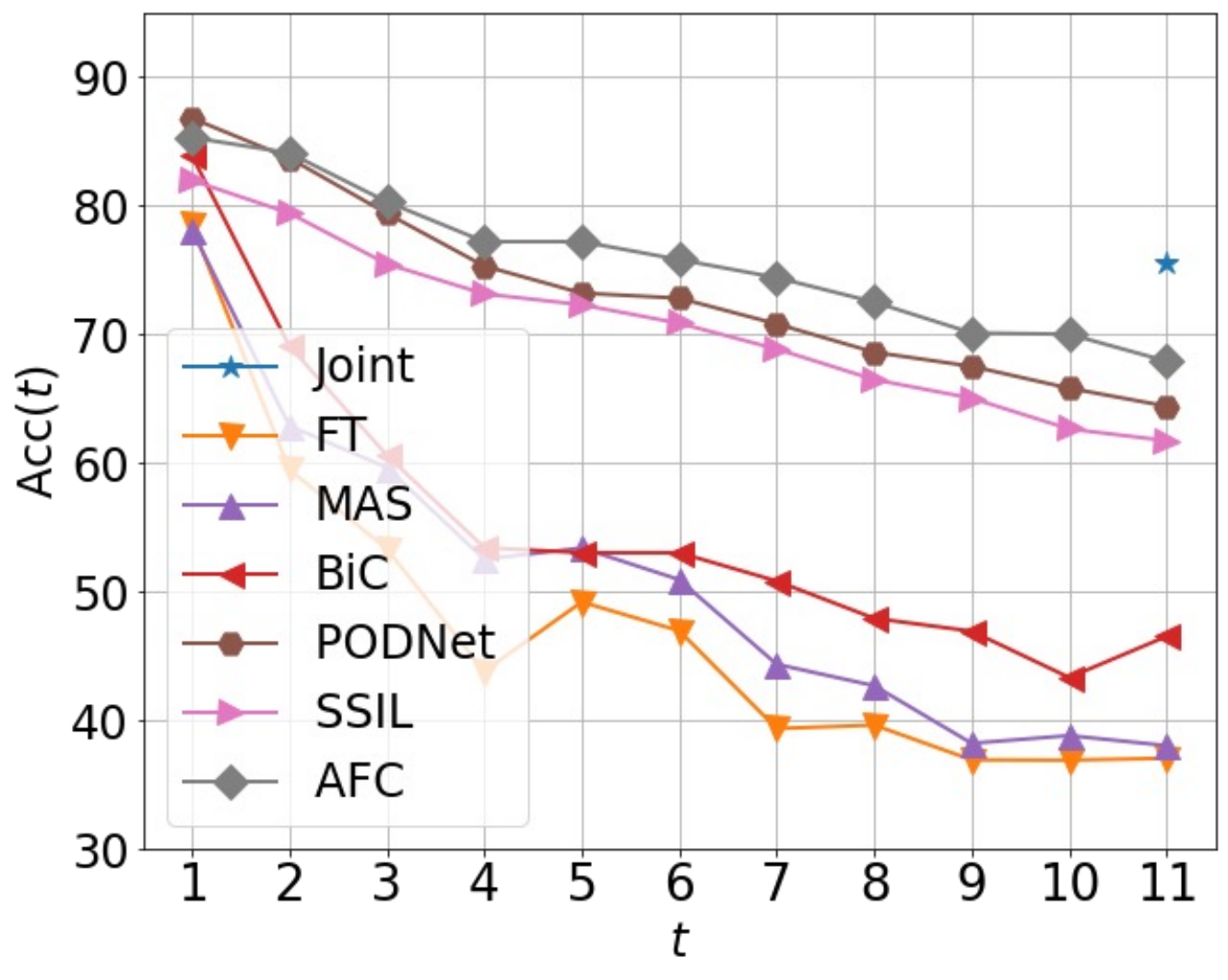}
\label{figure:11tasks_acc}}
\subfigure[$k$-NN($t$).]
{\includegraphics[width=0.32 \linewidth]{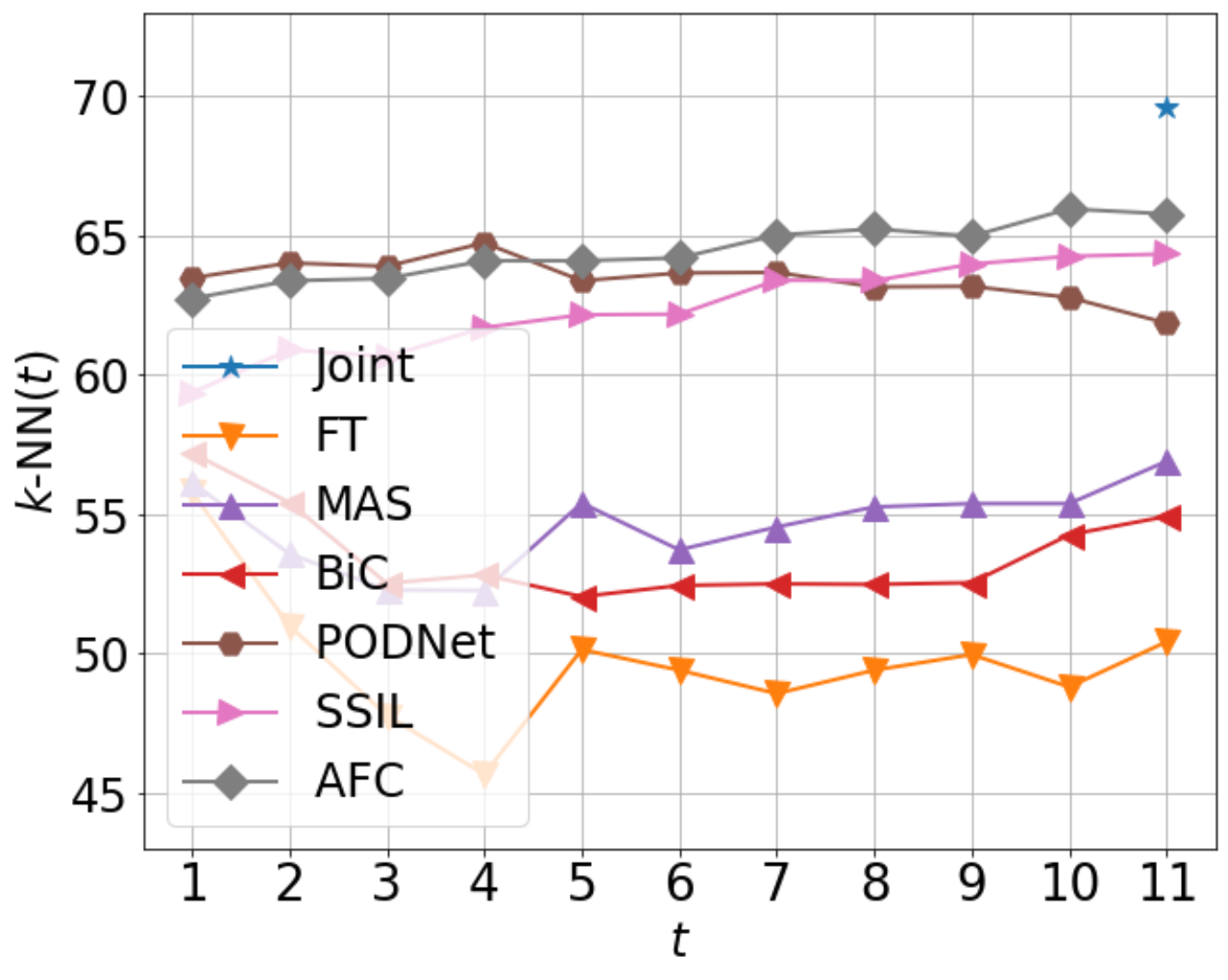}
\label{figure:11tasks_knn}}
\subfigure[All eval. metrics.]
{\includegraphics[width=0.32 \linewidth]{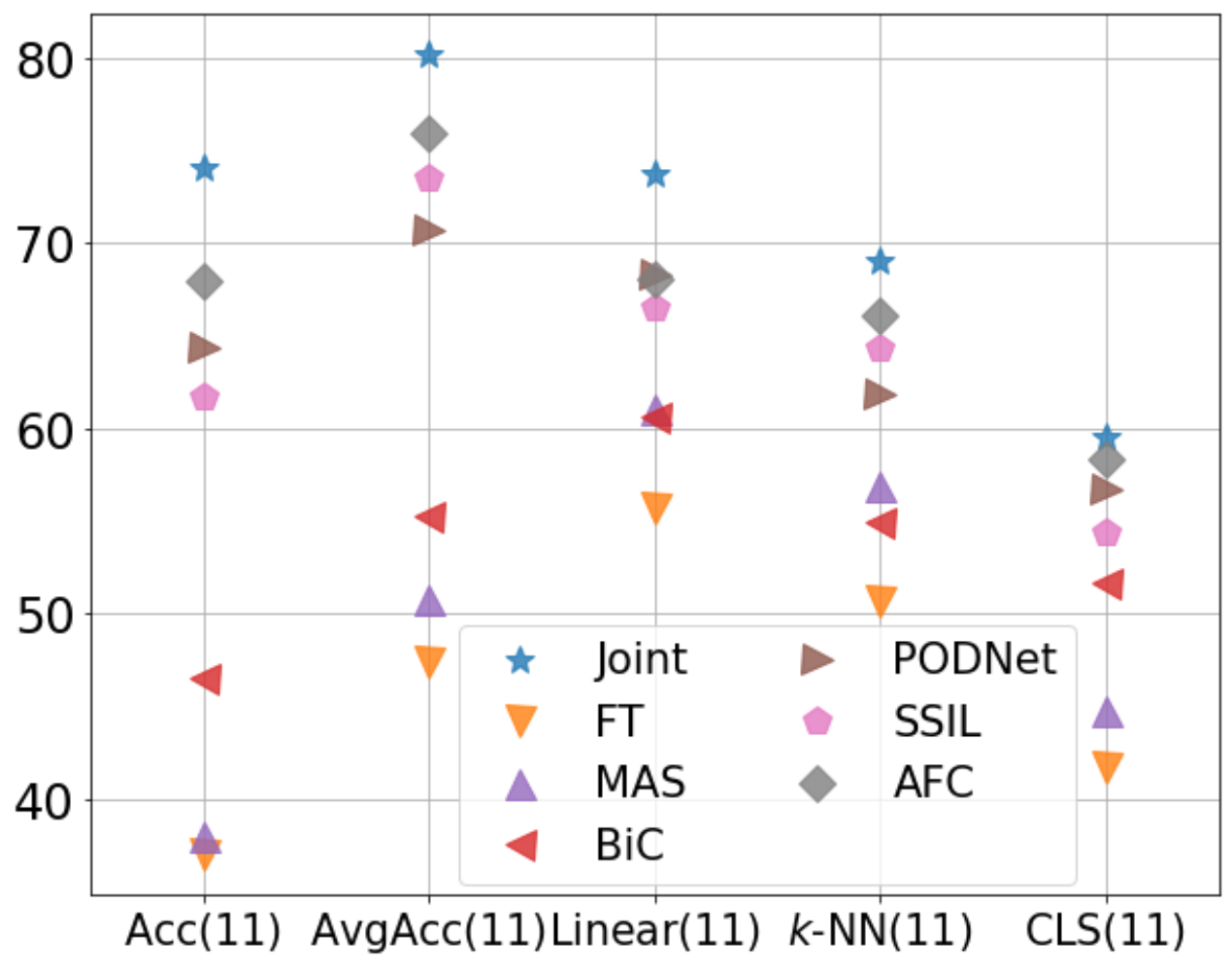}
\label{figure:11tasks_metrics}}
\vspace{-.1in}
\caption{{The experimental results of regularization-based CIL algorithms for a 11-task scenario using the ImageNet-100 dataset. "Joint" refers to the performance of the upper bound case using the entire datasets.}}
\vspace{-.1in}
\label{figure:11tasks_figures}
\end{figure*}

{For the ImageNet trained models, experimental evidence indicates that models achieving superior classification accuracy also learn a superior representation~\citep{(transferlearning)kornblith2019better}.
On the other hand, many regularization-based CIL methods devise novel regularization  motivated by the goal of learning new knowledge (\textit{i.e.}, representation) effectively while retaining knowledge from previous tasks. As a result, they demonstrate superior performance in the final classification accuracy~\citep{(cil_survey)masana2020class}.
Inspired by these findings, we ask the following question: Do regularization-based CIL algorithms using exemplar memory attain their excellence by learning a superior representation? To explore this, we construct experiments on 10-tasks and 11-tasks scenarios using the ImageNet-100 dataset and apply the proposed analysis after training models using selected regularization-based CIL algorithms.}

 {Figure \ref{figure:10tasks_figures} presents the experimental results for the 10-tasks scenario. First, Figure \ref{figure:10tasks_acc} shows CIL algorithm's \textbf{Acc}($t$). As reported in their paper, state-of-the-art algorithms (\textit{i.e.}, AFC, SSIL, and PODNet) demonstrate the most superior performance, with BiC following closely behind. 
 Additionally, not only MAS and BiC (w/o BC) achieve worse performance compared to them but also BiC (w/o BC) shows degraded performance as reported in their paper~\citep{(bic)wu2019large}.
Figure \ref{figure:10tasks_knn} and \ref{figure:10tasks_entire} show the result of applying our proposed evaluation.
From these results, we observe several interesting findings: First, from the $k$\textbf{-NN}($t$) results in Figure \ref{figure:10tasks_knn}, we can confirm that the results of conventional metrics do not always align with the quality of learned representations. 
For instance, in $k$\textbf{-NN}($t$), BiC and MAS show superior representations compared to the state-of-the-art algorithms. 
Specifically, considering that the representation of BiC is the same as BiC (w/o BC)~\citep{(bic)wu2019large}, CIL using only the knowledge distillation method (\textit{e.g.}, LWF) achieves significantly degraded performance in \textbf{Acc}($t$) due to biased prediction but demonstrates a better representation learning capability. 
Furthermore, the additional results in Figure \ref{figure:10tasks_entire} highlight these trends more distinctly. Despite the state-of-the-art algorithms achieving superior performance in the conventional metrics (\textit{i.e.}, \textbf{Acc(10)} and \textbf{AvgAcc(10)}), they exhibit the same trend in all metrics evaluating representation quality as before. 
Particularly, despite PODNet achieving relatively superior performance in the conventional metrics, the representation they learned are significantly worse than others.}

{In contrast, the results of the 11-tasks in Figure \ref{figure:11tasks_figures} exhibit a different trend. First, from the \textbf{Acc}($t$) in Figure \ref{figure:11tasks_acc}, we can observe that the state-of-the-art algorithms continue to demonstrate the most superior performance and are almost approaching performance of Joint. Unlike the 10-tasks scenario, the results in Figure \ref{figure:11tasks_knn} demonstrate that these state-of-the-art methods learn better representations than others. Particularly, AFC and SSIL show a slight improvement in quality of learned representations over tasks. Similarly, Figure \ref{figure:11tasks_metrics} demonstrate the state-of-the-art algorithms achieve not only  superior performance in the conventional metrics but also superior representation learning.}

{We have observed that several state-of-the-art CIL algorithms exhibit completely different trends in representation learning between the 10-tasks and 11-tasks scenarios. The only difference between these scenarios is that in the 11-tasks scenario, the models start by learning half of the classes from the first task. Taking this difference into consideration, we conduct additional analysis to understand the reasons behind these results and to gain further insights into characteristics of these algorithms.}

\begin{figure}[h]
\centering 
\subfigure[Joint]
{\includegraphics[width=0.24\linewidth]{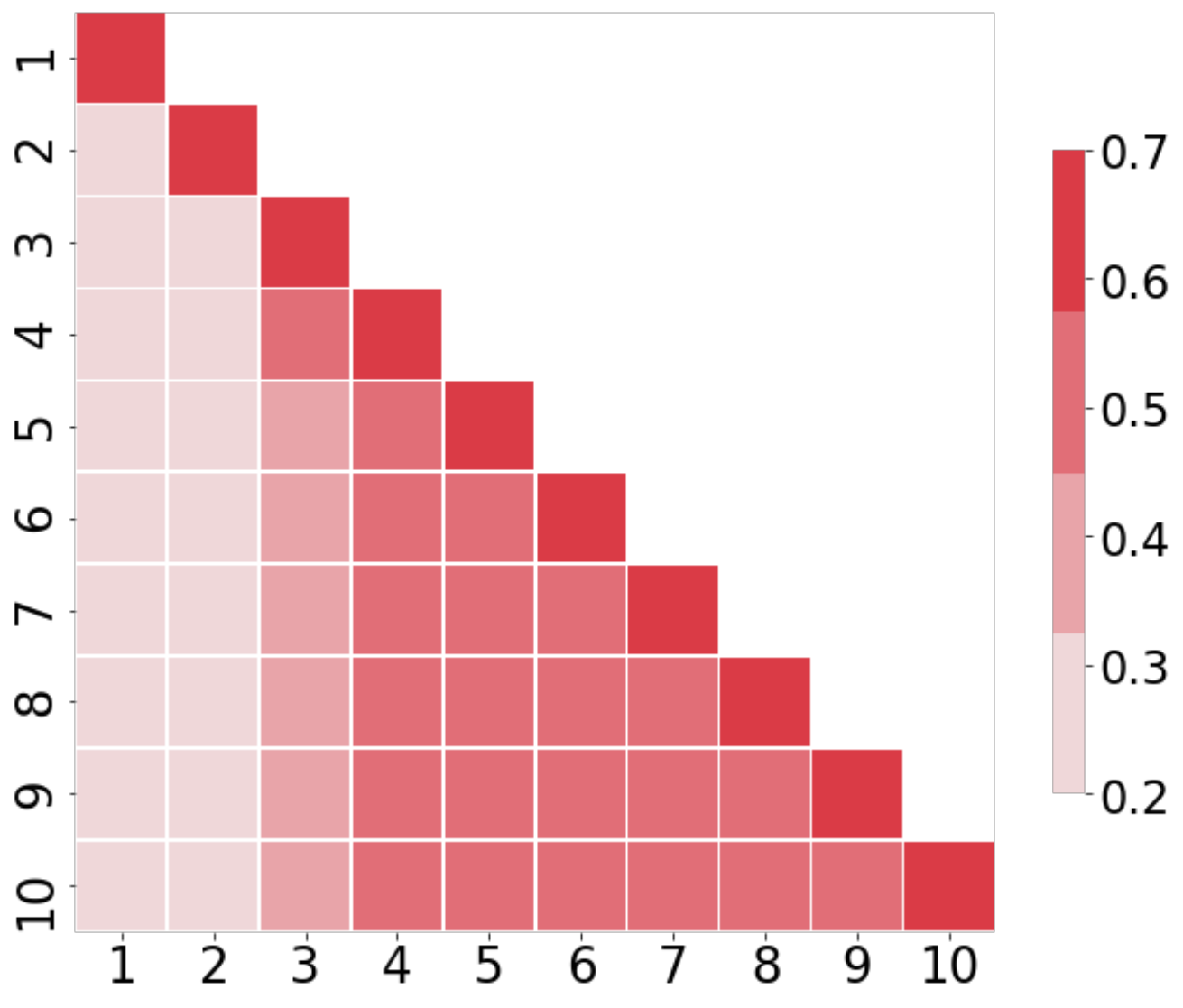}
\label{figure:10tasks_cka_joint}}
\subfigure[BiC]
{\includegraphics[width=0.24 \linewidth]{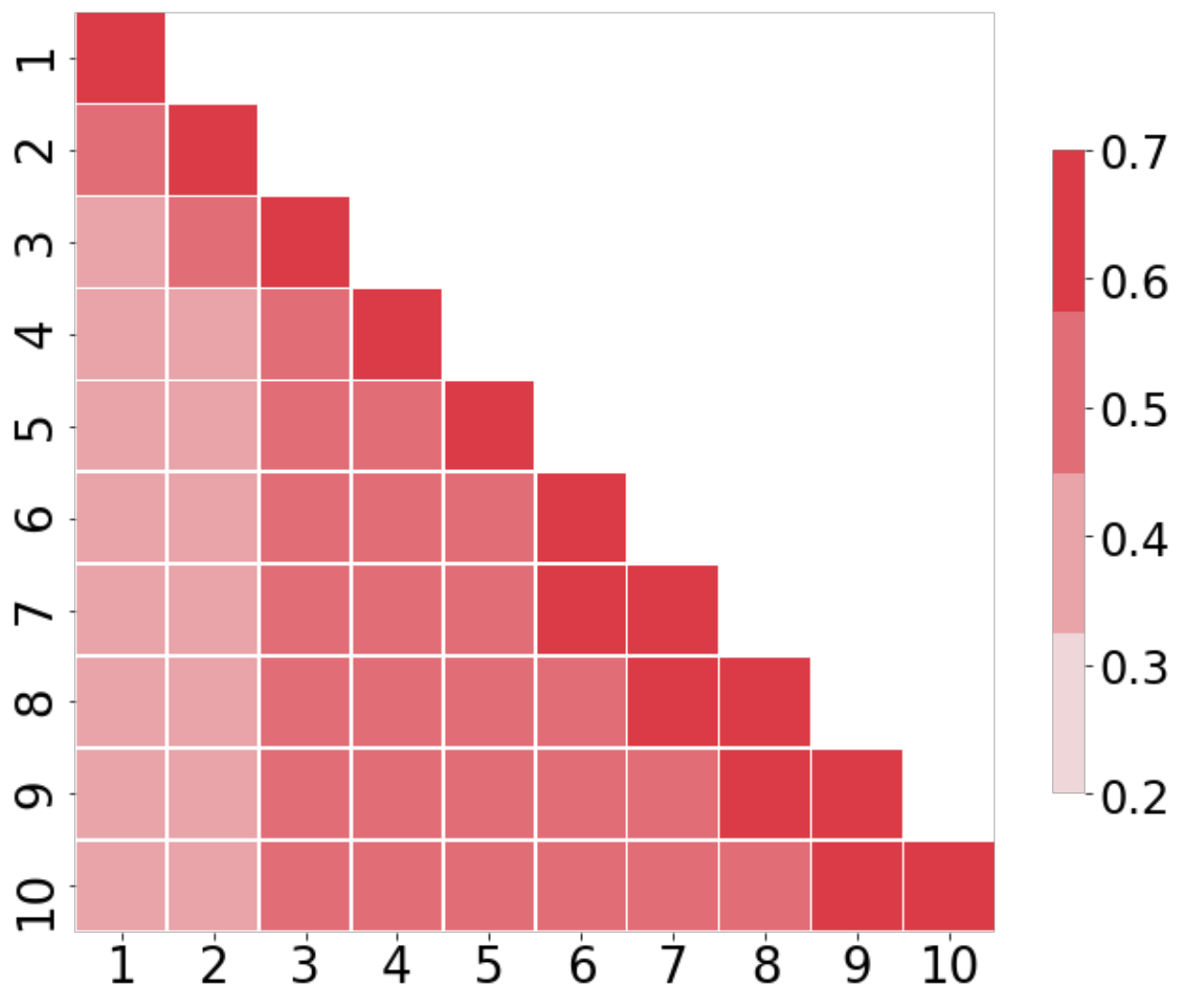}
\label{figure:10tasks_cka_bic}}
\subfigure[PODNet]
{\includegraphics[width=0.24\linewidth]{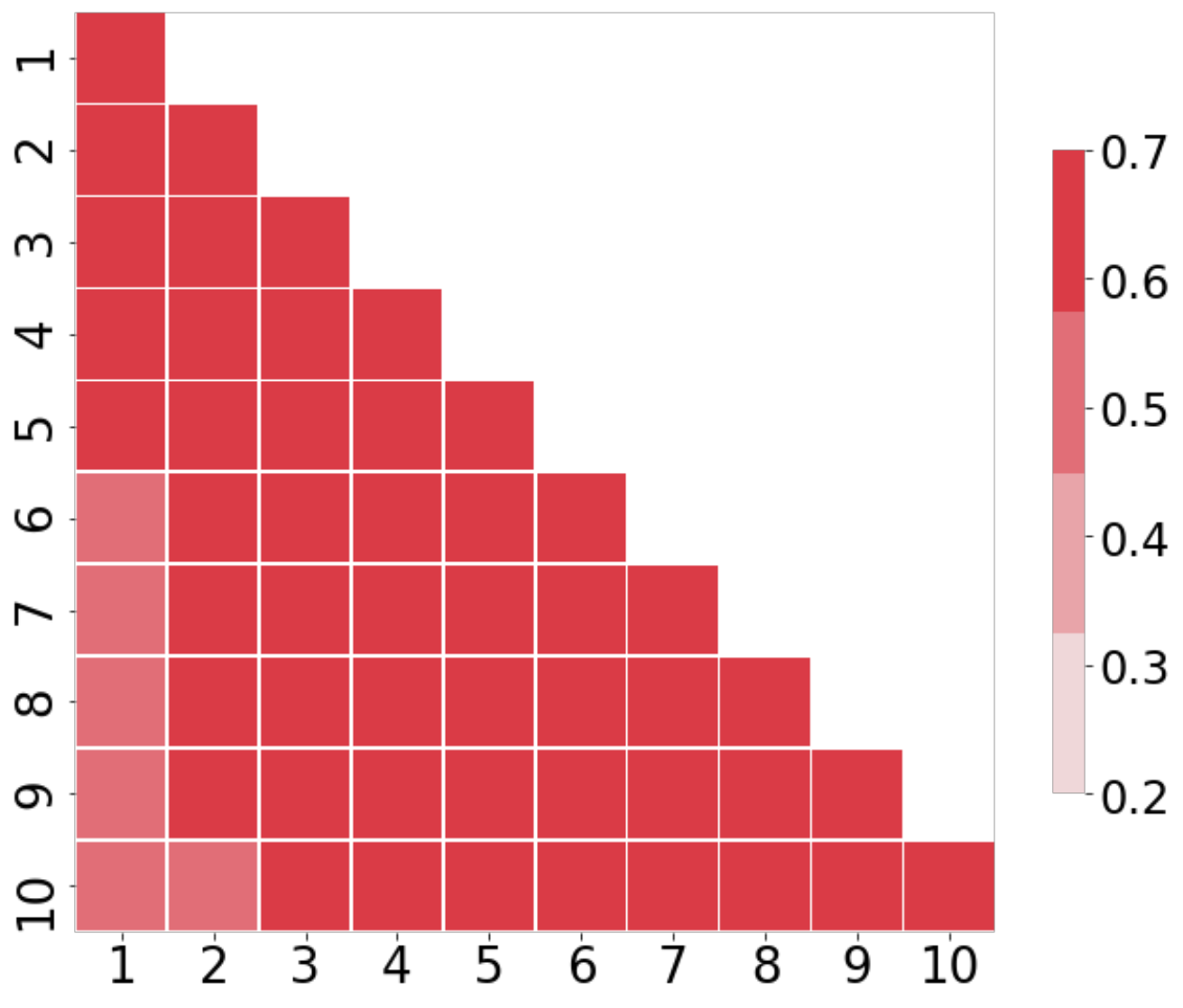}
\label{figure:10tasks_cka_podnet}}
\subfigure[AFC]
{\includegraphics[width=0.24 \linewidth]{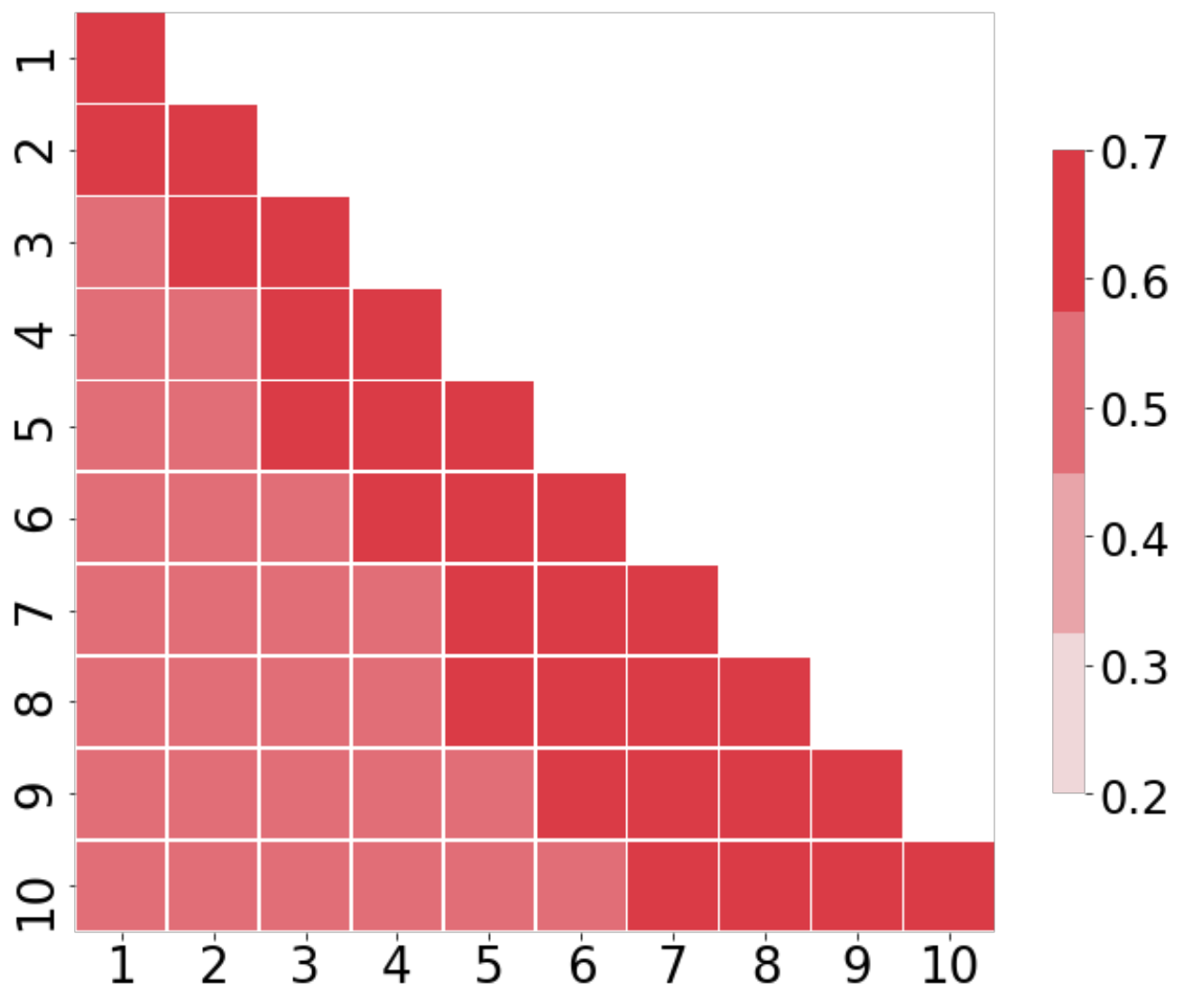}
\label{figure:10tasks_cka_afc}}
\caption{{{CKA}${(t_{1},t_{2})}$ in 10-tasks scenario for $t_{1}, t_{2}\in\{1,\dots,10\}$. Each ${CKA}(t_{1},t_{2})$ quantifies the similarity between representations of two models trained on distinct tasks. A deep red color indicates a higher level of similarity compared to a lighter shade of red.}}
\vspace{-.1in}
\label{figure:10tasks_cka}
 \end{figure}

\begin{figure}[h]
\centering 
\subfigure[Joint]
{\includegraphics[width=0.23\linewidth]{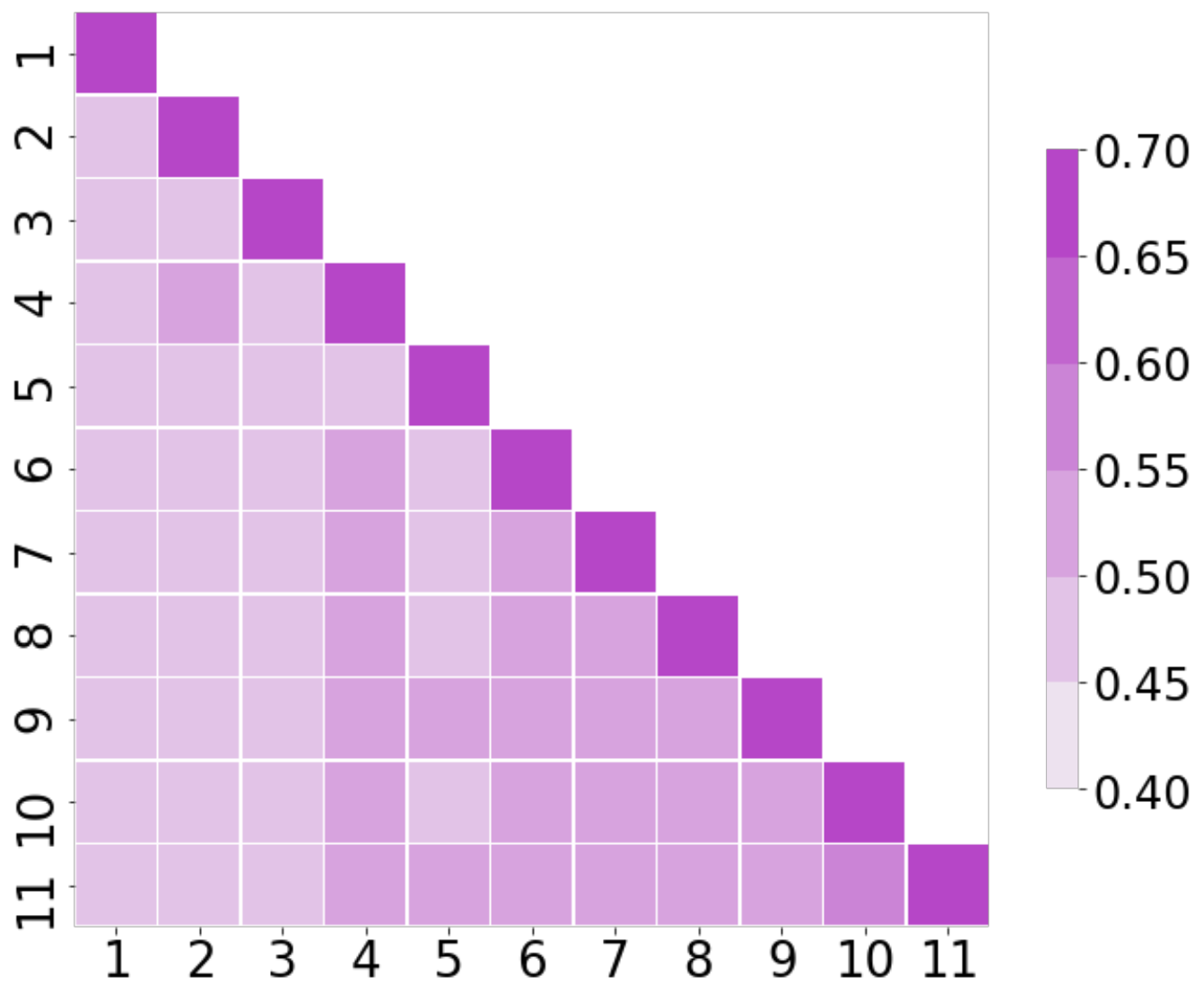}
\label{figure:11tasks_cka_joint}}
\subfigure[BiC]
{\includegraphics[width=0.23\linewidth]{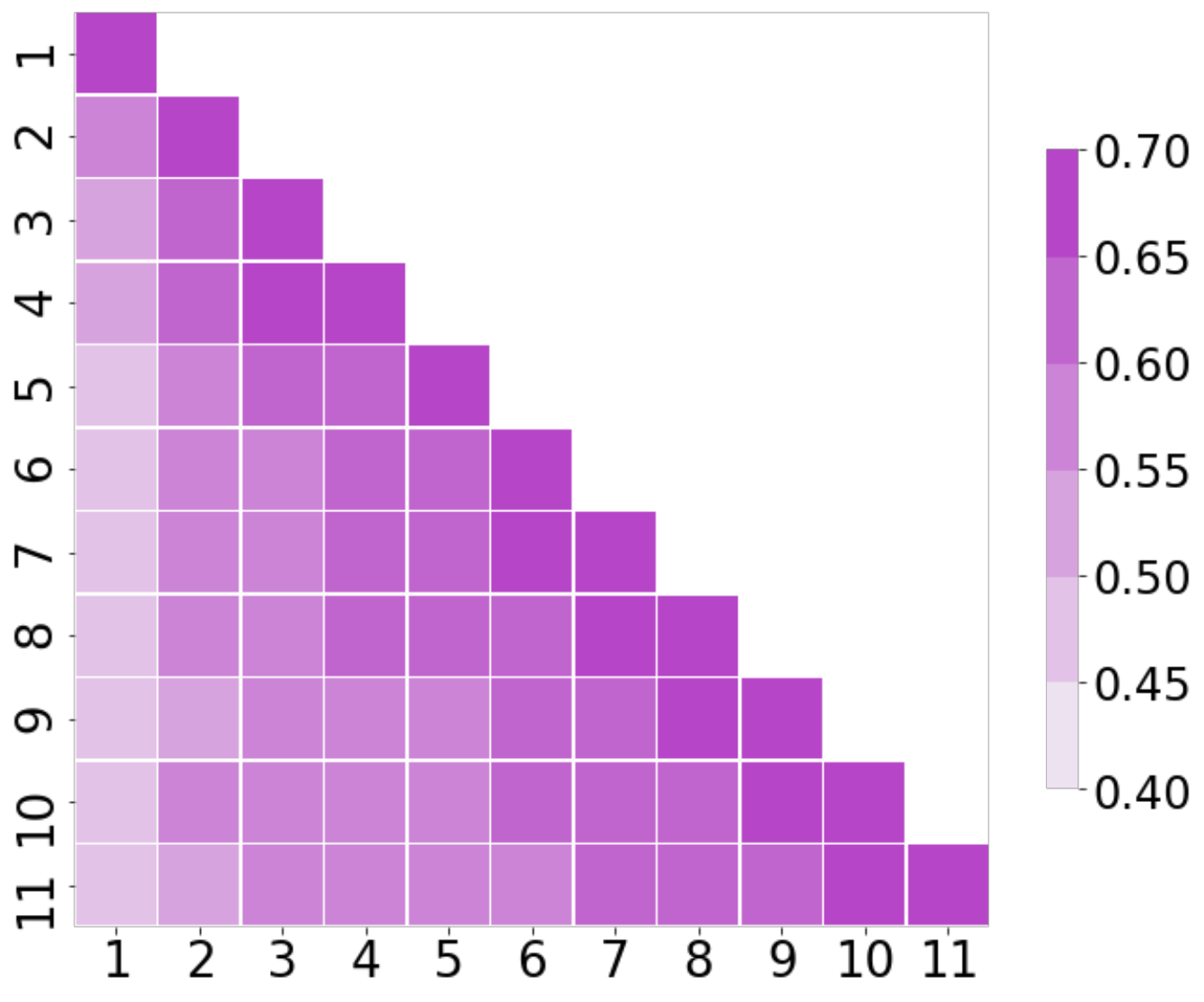}
\label{figure:11tasks_cka_ssil}}
\subfigure[PODNet]
{\includegraphics[width=0.23 \linewidth]{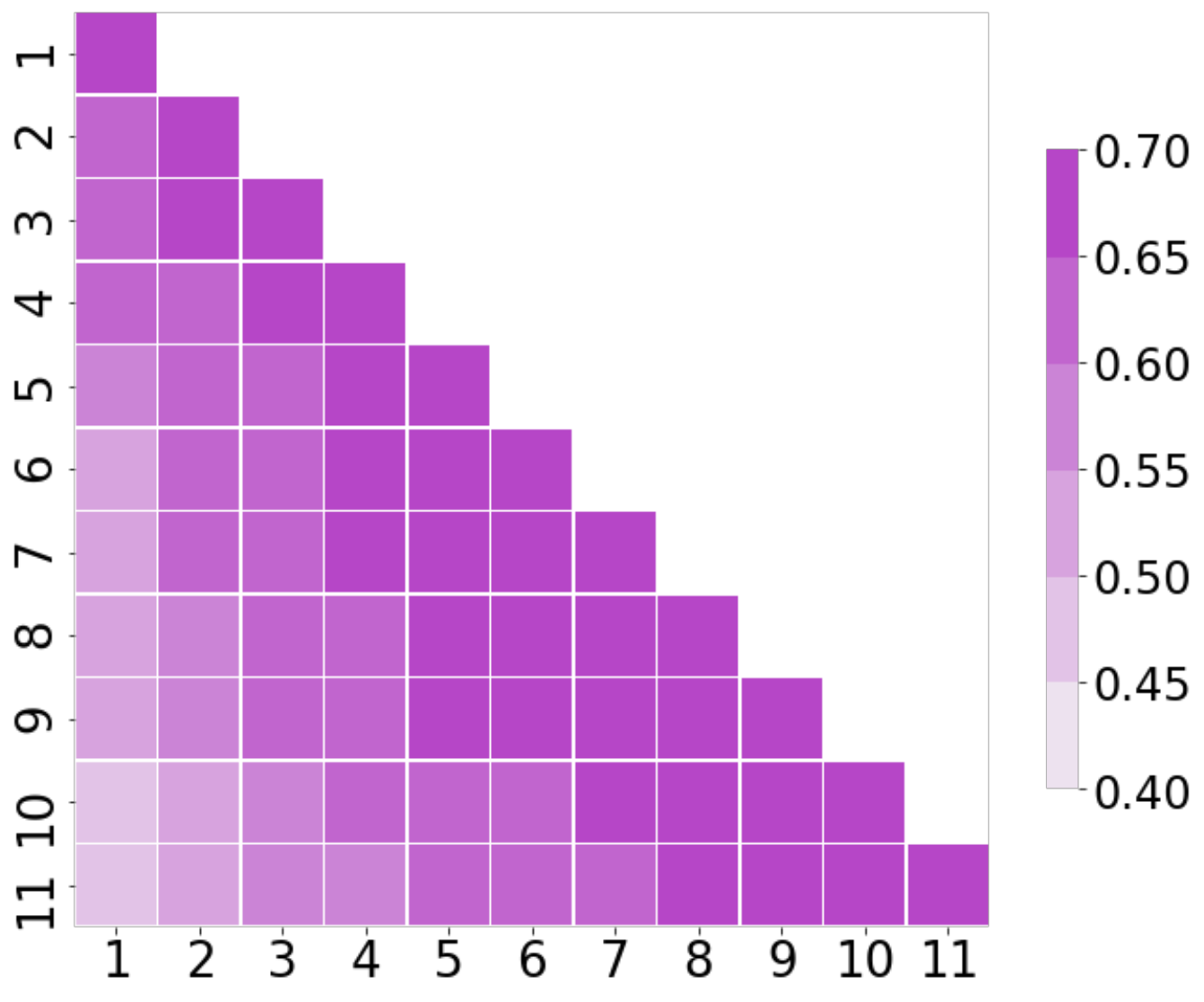}
\label{figure:11tasks_cka_podnet}}
\subfigure[AFC]
{\includegraphics[width=0.23 \linewidth]{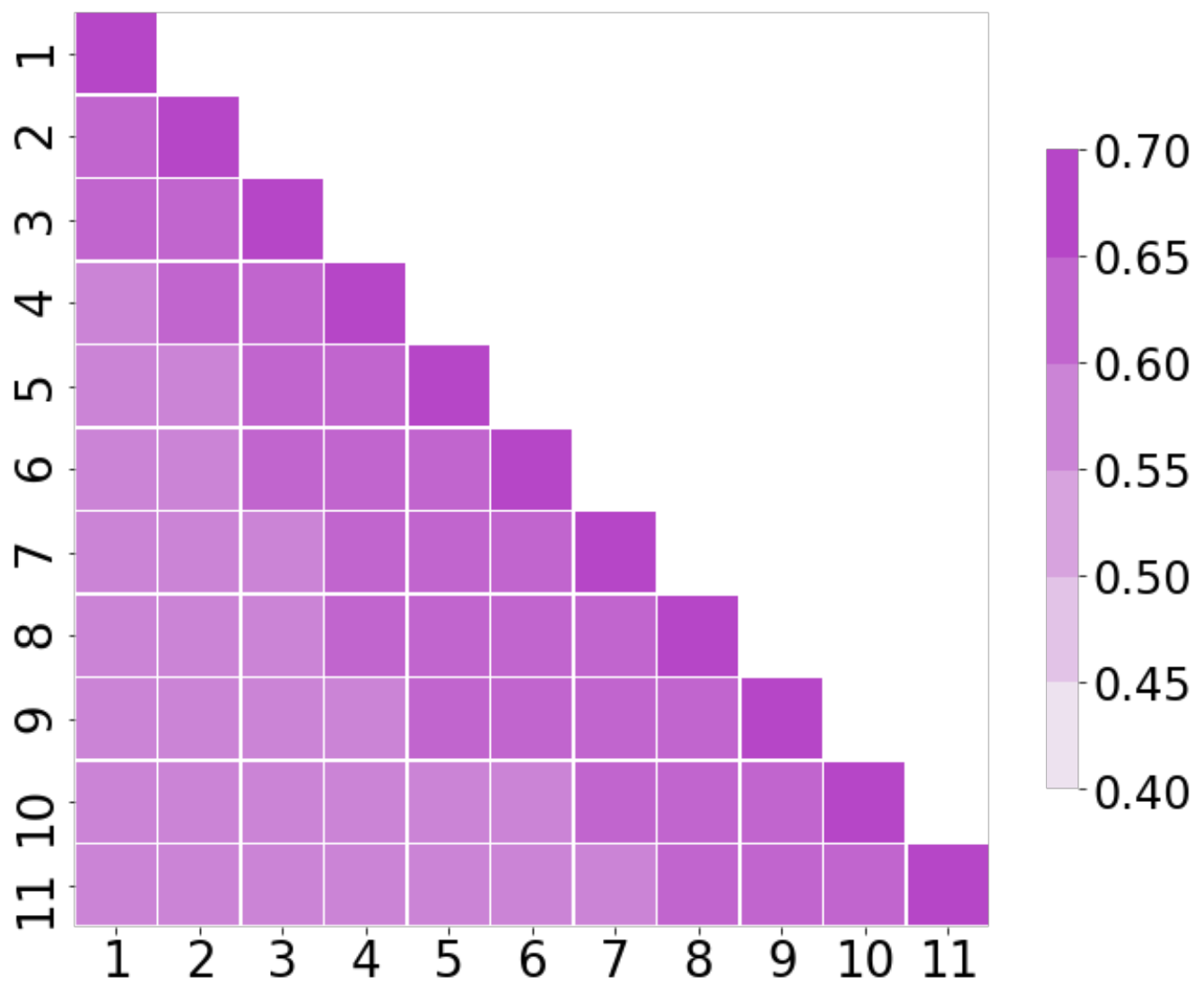}
\label{figure:11tasks_cka_afc}}
\vspace{-.1in}
\caption{{{CKA}${(t_{1},t_{2})}$ in 11-tasks scenario for $t_{1}, t_{2}\in\{1,\dots,11\}$. Each ${CKA}(t_{1},t_{2})$ quantifies the similarity between representations of two models trained on distinct tasks. A deep purple color indicates a higher level of similarity compared to a lighter shade of purple.}}
\vspace{-.1in}
\label{figure:11tasks_cka}
 \end{figure}

\subsection{{Most regularization-based CIL algorithms significantly prioritize stability }}

{In previous experimental results, we observed that state-of-the-art algorithm learn inferior representations in the 10-tasks scenario. However, in the 11-tasks scenario, while they learn superior representations, there is no significant difference when compared to the representations learned in the first task. This led us to hypothesize that the poor sequential representation updates are due to the strong stability of these algorithms. Consequently, we conduct additional analysis using CKA to investigate further.}

{Figure \ref{figure:10tasks_cka} and  \ref{figure:11tasks_cka} shows \textbf{CKA}${(t_{1},t_{2})}$ of Joint, BiC, PODNet, and AFC, which compares the representation similarity between $h_{\bm \psi_{t_{1}}}$ and $h_{\bm \psi_{t_{2}}}$. For example, \textbf{CKA}($5,1$) and \textbf{CKA}($5,3$) of Joint in Figure \ref{figure:10tasks_cka_joint} shows that an encoder at task $5$ has less representation similarity to the encoder at task $1$ compared to the encoder at task $3$. From the results of both the 10-tasks and 11-tasks scenarios, we can draw the following findings: First, the Joint exhibits a high similarity in representations between adjacent tasks, but ultimately, it undergoes progressive changes throughout the CIL process, resulting in superior learned representations at the final task. 
However, the representation similarity of PODNet and AFC remains relatively high. This indicates that these algorithms place a heavy emphasis on stability, leading to minimal changes in representations.}

{This aligns with the \textbf{$k$-NN} results in the previous section. 
In the 11-task scenario, the superior representation learned in the first task is consistently maintained, enabling these algorithms to achieve better results by the final task compared to other algorithms. Conversely, in the 10-tasks scenario, the representation learned in the first task is not as superior, preventing the algorithms from learning improved representations during the CIL process, which results in poorer performance compared to other baselines.}

\subsection{The superior performance of state-of-the-art algorithms might be attributed to their ability to learn a good output layer.}

{Additionally, one remaining question is how does the state-of-the-art algorithms (\textit{i.e.}, PODNet, SSIL, and AFC) can {still achieve high} \textbf{Acc}($t$) and \textbf{AvgAcc}($t$) even with \textit{poor representations}? To obtain an answer to this question, we conduct an analysis on weights of output layer.}
\begin{wrapfigure}{r}{0.5\textwidth}
\centering \includegraphics[width=0.98\linewidth]{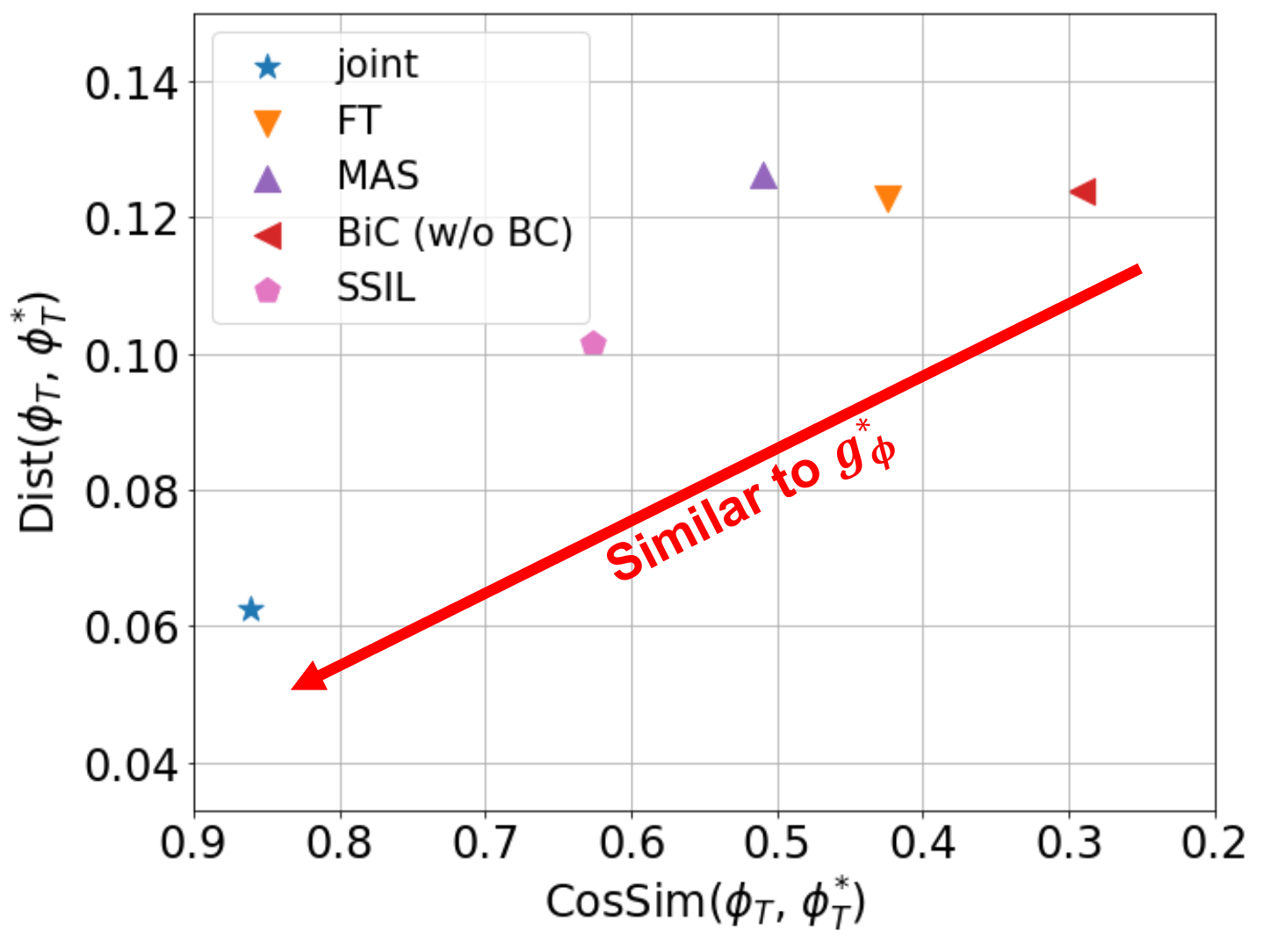}
\vspace{-.1in}
\caption{Experimental results of {CosSim}$(\phi_{T}, \phi^{*}_{T})$ and {Dist}$(\phi_{T}, \phi^{*}_{T})$. We use a model trained by each algorithm in the 10-tasks scenario.}
\vspace{-.1in}
\label{figure:output_layer}
\end{wrapfigure}
 {Figure \ref{figure:output_layer} compares the similarity of weights between an original classifier learned in last task (\textit{i.e.}, $g_{\bm \phi_T}$) and the estimated linear classifier trained for linear probing (\textit{i.e.}, $g^{*}_{\bm \phi_T}$). Note that results for PODNet and AFC are absent, as these algorithms adopt specialized cosine classifiers to address biased prediction in output layer. }
 {First, Joint demonstrates both the highest cosine similarity (\textbf{CosSim}$(\phi_{T}, \phi^{}{T})$) and lowest $L_2$ distance (\textbf{Dist}$(\phi{T}, \phi^{}_{T})$). Conversely, BiC(w/o BC) exhibits the lowest similarity to the estimated linear classifier. Furthermore, SSIL achieves a higher similarity to the estimated linear classifier compared to FT and MAS, benefiting from its novel approach which alleviates the biased prediction. Note that the order of similarity of output layers mirrors the performance order in Figure \ref{figure:10tasks_acc}. 
 Considering this trend alongside the experimental results on representation quality in Figure \ref{figure:10tasks_entire}, it is evident that SSIL learns representations comparable to those of FT and MAS, and inferior to BiC. Despite this, SSIL achieves overwhelmingly superior performance in conventional metrics, significantly benefiting from learning an effective output layer rather than from learning superior representations.}

{Similarly, we can draw a comparable inference for other state-of-the-art algorithms such as PODNet and AFC. 
Even though these algorithms learn representations that are inferior or not significantly different from those of other baselines, they can still achieve superior performance in conventional metrics thanks to the use of a specialized cosine classifier. 
This indicates that their performance advantage likely stems not from their novel regularization devised for controlling a trade-off between stability and plasticity in representations, but rather from the sophisticated design of the specialized output layer they commonly employ.}
 
\begin{figure}[h]
\centering 
\subfigure[10-tasks]
{\includegraphics[width=0.48\linewidth]{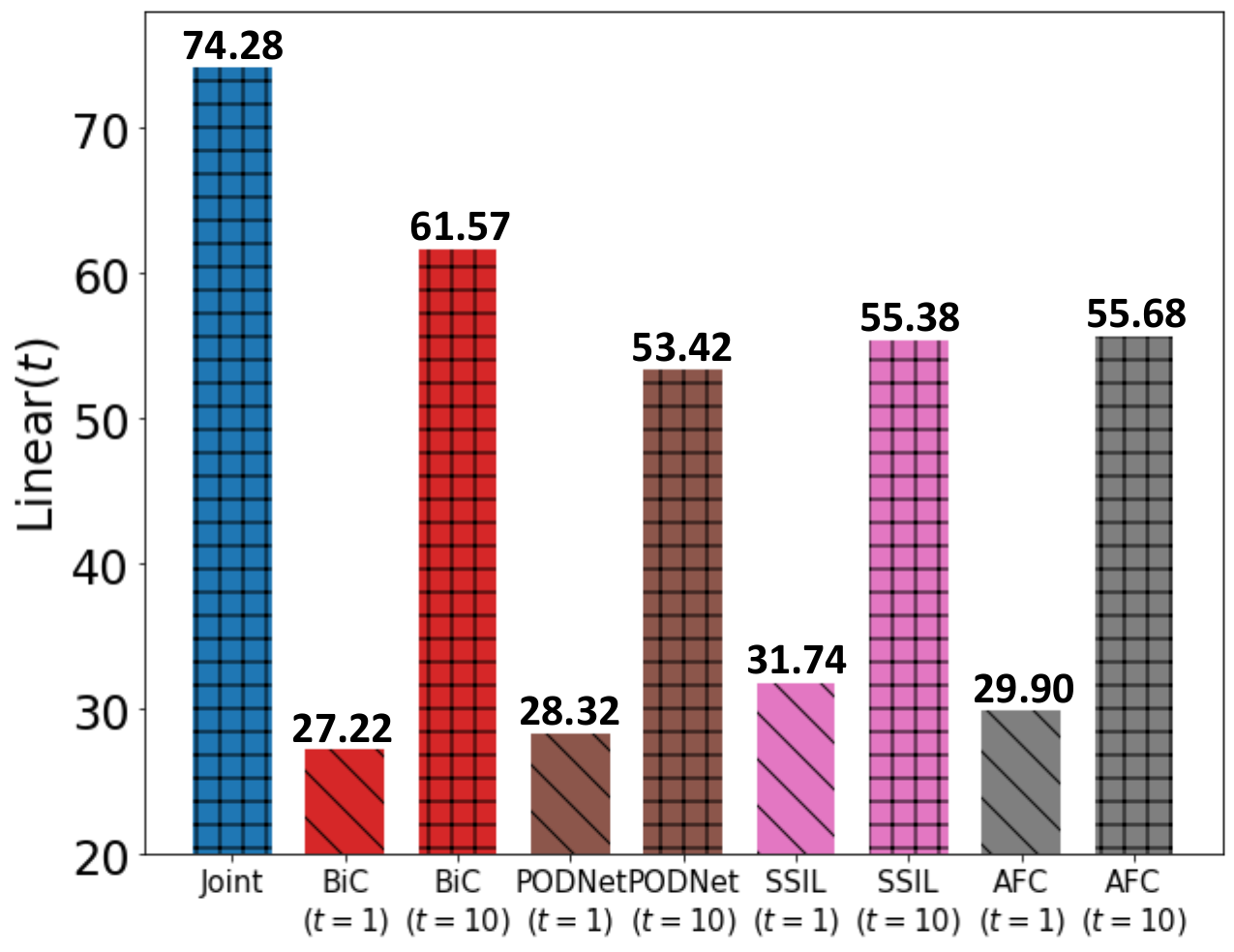}
\label{figure:base_10tasks}}
\subfigure[11-tasks with the base task]
{\includegraphics[width=0.48\linewidth]{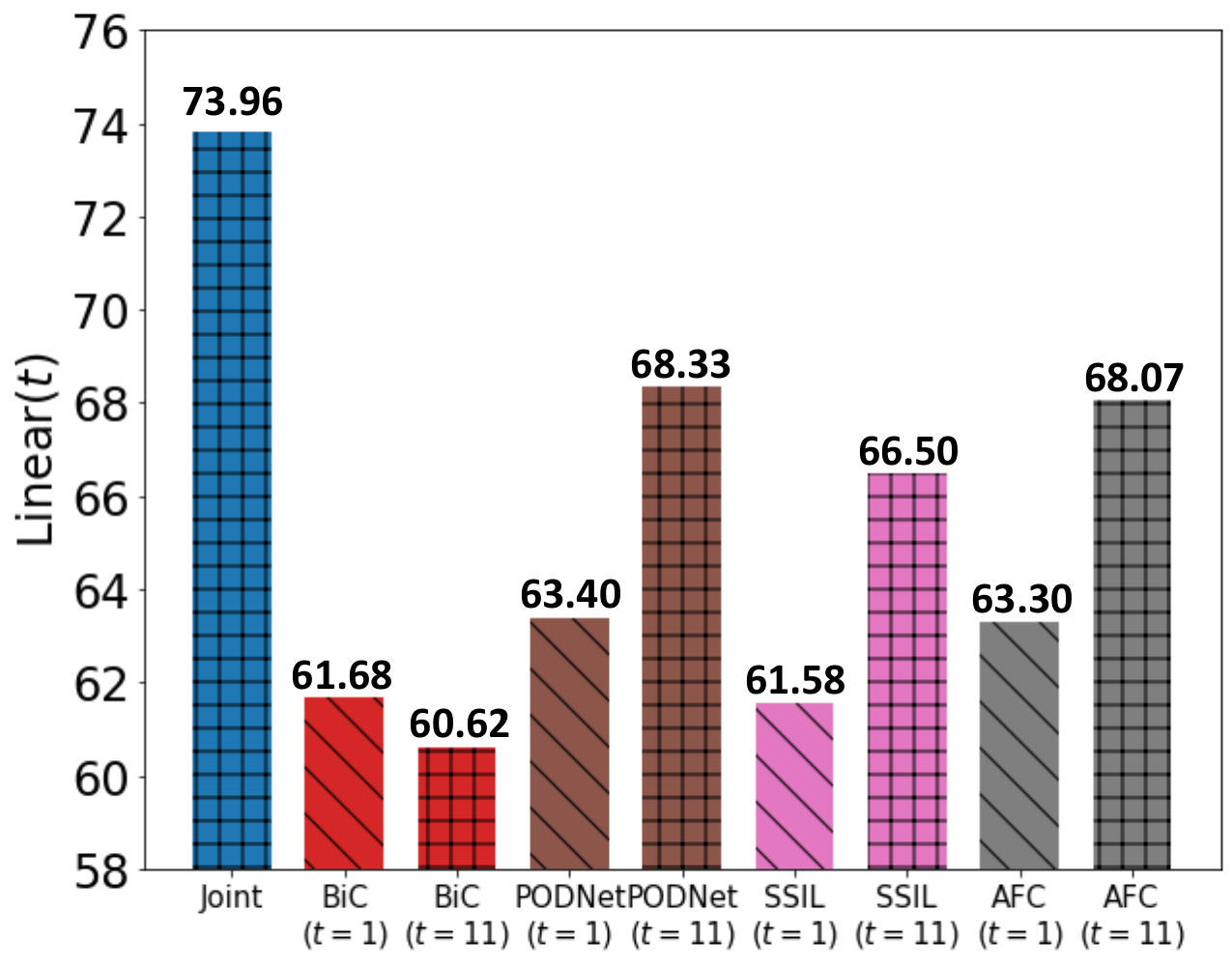}
\label{figure:base_12tasks}}
\vspace{-.1in}
\caption{{Experimental results of linear probing for each algorithm.}}
\vspace{-.15in}
\label{figure:base_tasks}
\end{figure}

\subsection{{The quality of the representation learned in the first task can have a significant impact on the final evaluation}}
{In the previous section, we noted significant differences in the evaluation results of the first task models among various algorithms, as illustrated in Figures \ref{figure:10tasks_knn} and \ref{figure:11tasks_knn} (e.g., \textbf{$k$-NN}(1)). We recognize that learning the initial task in CIL resembles single-task learning and is mostly unaffected by a particular regularization-based CIL algorithm. Considering the focus of these CIL algorithms on stability, we hypothesize that the disparities among these first task models could significantly impact overall performance. To validate this hypothesis, we conduct additional experiments.}

{We compare the linear probing results of the first task model trained by each algorithm in the 10-tasks and 11-tasks scenarios, as shown in Figure \ref{figure:base_tasks}.
From this figure, despite employing the same dataset in the same scenario, we again observe that the linear probing results of the first task model exhibit variance across algorithms, as evidenced by the distinct \textbf{Linear}($1$) outcomes in the figure. }
{We specifically notice a consistent pattern in the performance discrepancies and ranking between \textbf{Linear}($1$) and \textbf{Linear}($T$) across all algorithms, as observed in the results of PODNet, SSIL, and AFC. For instance, in the 11-task scenario, the performance gap between PODNet and SSIL in \textbf{Linear}(1) is approximately 2\%, mirroring their difference in \textbf{Linear}($11$). This observation reinforces the notion that variations in learning the first task can indeed affect the final performance assessment, leading us to conduct experiments focused on aligning the first task model as closely as possible.}

\begin{table*}[!h]
\caption{Experimental results for the 10- and 11-tasks scenarios.}
\vspace{-.1in}
\label{table:entire_result}
\centering
\smallskip\noindent
\resizebox{0.98\linewidth}{!}{
\begin{tabular}{|c||cccccc||cccccc|}
\hline
\multirow{3}{*}{Alg.} & \multicolumn{6}{c||}{10-tasks}                                                                                                                                                                             & \multicolumn{6}{|c|}{11-tasks}                                                                                                                                                                            \\ \cline{2-13} 
                      & \multicolumn{2}{c|}{Conv. Metrics}                                      & \multicolumn{3}{c|}{In-domain }                                                                            & Out-domain  & \multicolumn{2}{c|}{Conv. Metrics}                                      & \multicolumn{3}{c|}{In-domain }                                                                           & Out-domain \\ \cline{2-13} 
                      & \multicolumn{1}{c|}{Acc($10$)}      & \multicolumn{1}{c|}{Avg($10$)}      & \multicolumn{1}{c|}{\textbf{Linear}($1$)}    & \multicolumn{1}{c|}{Linear($10$)}   & \multicolumn{1}{c|}{$k$-NN($10$)}   & CLS($10$)        & \multicolumn{1}{c|}{Acc($11$)}      & \multicolumn{1}{c|}{Avg($11$)}      & \multicolumn{1}{c|}{\textbf{Linear}($1$)}   & \multicolumn{1}{c|}{Linear($11$)}   & \multicolumn{1}{c|}{$k$-NN($11$)}   & CLS($11$)        \\ \hline \hline
Joint                 & \multicolumn{1}{c|}{75.56}          & \multicolumn{1}{c|}{80.23}          & \multicolumn{1}{c|}{-}              & \multicolumn{1}{c|}{74.28}          & \multicolumn{1}{c|}{69.62}          & 59.57            & \multicolumn{1}{c|}{75.40}          & \multicolumn{1}{c|}{84.23}          & \multicolumn{1}{c|}{-}             & \multicolumn{1}{c|}{73.96}          & \multicolumn{1}{c|}{74.42}          & 59.80            \\ \hline
FT                    & \multicolumn{1}{c|}{37.20}          & \multicolumn{1}{c|}{53.48}          & \multicolumn{1}{c|}{\textbf{33.12}}          & \multicolumn{1}{c|}{56.08}          & \multicolumn{1}{c|}{52.04}          & 44.09            & \multicolumn{1}{c|}{37.01}          & \multicolumn{1}{c|}{47.35}          & \multicolumn{1}{c|}{\textbf{61.22}}         & \multicolumn{1}{c|}{56.12}          & \multicolumn{1}{c|}{56.12}          & 41.72            \\ 
MAS                   & \multicolumn{1}{c|}{36.48}          & \multicolumn{1}{c|}{53.55}          & \multicolumn{1}{c|}{\textbf{32.66}}          & \multicolumn{1}{c|}{58.94}          & \multicolumn{1}{c|}{53.38}          & 48.91            & \multicolumn{1}{c|}{38.00}          & \multicolumn{1}{c|}{50.85}          & \multicolumn{1}{c|}{\textbf{60.62}}         & \multicolumn{1}{c|}{61.08}          & \multicolumn{1}{c|}{61.08}          & 44.79            \\ 
BiC                   & \multicolumn{1}{c|}{44.60}          & \multicolumn{1}{c|}{59.28}          & \multicolumn{1}{c|}{\textbf{27.22}}          & \multicolumn{1}{c|}{{61.58}} & \multicolumn{1}{c|}{{56.36}} & {52.20}   & \multicolumn{1}{c|}{46.50}          & \multicolumn{1}{c|}{55.31}          & \multicolumn{1}{c|}{\textbf{61.67}}         & \multicolumn{1}{c|}{60.61}          & \multicolumn{1}{c|}{63.40}          & 51.71            \\ \hline
PODNet                & \multicolumn{1}{c|}{50.70}          & \multicolumn{1}{c|}{66.70}          & \multicolumn{1}{c|}{\textbf{28.32}}          & \multicolumn{1}{c|}{53.42}          & \multicolumn{1}{c|}{40.26}          & 45.09            & \multicolumn{1}{c|}{64.40}          & \multicolumn{1}{c|}{73.48}          & \multicolumn{1}{c|}{\textbf{63.40}}         & \multicolumn{1}{c|}{{68.33}}          & \multicolumn{1}{c|}{{61.86}} & 56.67            \\ 
SSIL                  & \multicolumn{1}{c|}{53.14}          & \multicolumn{1}{c|}{67.12}          & \multicolumn{1}{c|}{\textbf{31.74}} & \multicolumn{1}{c|}{55.38}          & \multicolumn{1}{c|}{50.98}          & 47.28            & \multicolumn{1}{c|}{61.72}          & \multicolumn{1}{c|}{70.41}          & \multicolumn{1}{c|}{{\textbf{61.58}}}         & \multicolumn{1}{c|}{66.50}          & \multicolumn{1}{c|}{61.48}          & 54.39            \\ 
AFC                   & \multicolumn{1}{c|}{{54.90}} & \multicolumn{1}{c|}{{68.83}} & \multicolumn{1}{c|}{\textbf{29.90}}          & \multicolumn{1}{c|}{55.68}          & \multicolumn{1}{c|}{49.33}          & 46.39            & \multicolumn{1}{c|}{{67.90}} & \multicolumn{1}{c|}{{75.89}} & \multicolumn{1}{c|}{\textbf{63.30}} & \multicolumn{1}{c|}{{68.07}} & \multicolumn{1}{c|}{63.30}          & {58.38}   \\ \hline
\end{tabular}
}
\end{table*}

{Table \ref{table:entire_result} presents the results of conventional metrics and all proposed evaluations obtained from the 10-tasks and 11-tasks and we again confirm that \textbf{Linear}(1) of each algorithm is different. Our objective is to equalize the performance of \textbf{Linear}(1). For this, we set the same first task model with the lowest performance in each scenario for all algorithms.
For example, in the case of 10-tasks, we train a first task model with different epochs to obtain a model whose \textbf{Linear}(1) result closely matches 28.32 (the \textbf{Linear}(1) result of PODNet). Once we obtain a first task model with the similar result, we proceed with CIL using the original hyperparameters. The experimental results for this approach are shown in Table \ref{table:base}.
Instances marked with an asterisk (*) indicate results obtained using the unified first task model. 
Experimental results in the figure show that employing the unified first task model led to a decrease in performance across both scenarios. In the 10-tasks scenario, SSIL* and AFC* show performance decreases of 1-2\% in \textbf{Acc}(10) compared to their original results. Notably, for SSIL*, the difference in \textbf{Acc}(10) compared to PODNet reduces from roughly 4\% to around 0.6\%. Similarly, in the 11-tasks scenario, PODNet* and AFC* suffer from a 2-4\% decrease compared to the original results. As a result, although PODNet initially exhibits about a 3\% higher \textbf{Acc}(11) than SSIL, PODNet* ultimately performs worse than SSIL in terms of \textbf{Acc}(11).}


\begin{table}[!h]
\caption{Experimental results with a unified first task model. * denotes using the unified first task model.}
\label{table:base}
\centering
\smallskip\noindent
\resizebox{0.6\linewidth}{!}{
\begin{tabular}{|c||cccccc|}
\hline
\multirow{2}{*}{Alg.} & \multicolumn{6}{c|}{10-tasks}                                                                                                                                                     \\ \cline{2-7} 
                      & \multicolumn{1}{c|}{Acc($T$)} & \multicolumn{1}{c|}{Avg($10$)} & \multicolumn{1}{c|}{\textbf{Linear}($1$)} & \multicolumn{1}{c|}{Linear($10$)} & \multicolumn{1}{c|}{$k$-NN($10$)} & CLS($10$) \\ \hline
SSIL*                  & \multicolumn{1}{c|}{51.34}    & \multicolumn{1}{c|}{65.74}    & \multicolumn{1}{c|}{\textbf{28.48}}       & \multicolumn{1}{c|}{54.54}       & \multicolumn{1}{c|}{52.26}       & 45.28    \\ \hline
AFC*                   & \multicolumn{1}{c|}{53.70}    & \multicolumn{1}{c|}{67.85}    & \multicolumn{1}{c|}{\textbf{28.48}}       & \multicolumn{1}{c|}{51.02}       & \multicolumn{1}{c|}{46.82}       & 43.28    \\ \hline \hline
\multirow{2}{*}{Alg.} & \multicolumn{6}{c|}{11-tasks}                                                                                                                                \\ \cline{2-7} 
                      & \multicolumn{1}{c|}{Acc($11$)} & \multicolumn{1}{c|}{Avg($11$)} & \multicolumn{1}{c|}{\textbf{Linear}($1$)} & \multicolumn{1}{c|}{Linear($11$)} & \multicolumn{1}{c|}{$k$-NN($11$)} & CLS($11$) \\ \hline
PODNet*                & \multicolumn{1}{c|}{60.90}    & \multicolumn{1}{c|}{70.11}    & \multicolumn{1}{c|}{\textbf{61.28}}       & \multicolumn{1}{c|}{63.28}       & \multicolumn{1}{c|}{55.26}       & 51.67    \\ \hline
AFC*                   & \multicolumn{1}{c|}{65.90}    & \multicolumn{1}{c|}{74.48}    & \multicolumn{1}{c|}{\textbf{61.24}}       & \multicolumn{1}{c|}{64.34}       & \multicolumn{1}{c|}{61.70}       & 56.80    \\ \hline
\end{tabular}
}
\end{table}



{Additionally, comparing Table \ref{table:entire_result} and Table \ref{table:base} highlights shifts in representation quality when we establish the unified first task model. This is particularly noticeable in the 11-tasks scenario, where adjustments in the first task model significantly affect the representation quality of the final model. For instance, both PODNet and AFC begin learning from a model with approximately a 2\% decrease in \textbf{Linear}(1). This consequently results in notable performance declines of around 4\% at \textbf{Linear}(11) and 2-6\% at \textbf{$k$-\textbf{NN}}(11) and \textbf{CLS}(11).}


In this section, we propose novel findings through evaluating the representations of the first task model learned by each CIL algorithm. Note that, during the hyperparameter tuning process, most algorithms select the best hyperparameters that achieve the highest \textbf{Acc} or \textbf{AvgAcc} after learning the final task. Considering that most state-of-the-art regularization-based CIL algorithms heavily prioritize stability, it is plausible that the optimal hyperparameters, which can learn the best representation in the first task and preserve it in subsequent tasks, could be chosen as the best  hyperparameters, especially in scenarios where many classes are learned in the first task (\textit{e.g.}, 11-tasks). Indeed, when we equalized the first task model, we observed a significant reduction in the differences in final performance across algorithms. This suggests the importance of evaluating the quality of model representations in each task to accurately assess the performance gains of each CIL algorithm.

\section{Concluding Remarks, Limitation and Future Work}
 \noindent{\textbf{Key insights from experiments}} \  \
Based on experiments with state-of-the-art regularization-based CIL algorithms, we first confirm that evaluation results based on conventional metrics not align with the evaluation results of representations learned by each algorithm. Second, due to the heavy emphasis on stability in most state-of-the-art regularization-based CIL algorithms, representations do not change significantly during the CIL process. As a result, these algorithms achieve significant advantages primarily in scenarios where the first task involves learning many classes. Third, we demonstrate that the performance gains of these algorithms may stem more from a sophisticated output layer than from novel regularization terms. Lastly, we note that the representations learned in the first task can vary significantly across algorithms, significantly impacting the final evaluation.

 \noindent{\textbf{The importance  of evaluation in a representation learning perspective}} \  \
Neglecting the assessment of representations can lead to favorably evaluating algorithms that learn a good output layer despite having poorer representation quality. This not only inaccurately evaluates CIL algorithms but also limits their maximum potential to subpar representations instead of achieving the representation of the joint model. Building on these findings, We question whether the prevailing evaluation method, focused solely on classification accuracy, truly captures the factors that drive the performance improvement of each algorithm. In this regard, we highlight the need for diverse forms of evaluation, especially from a representation learning perspective, to precisely understand the performance gain of them.

 \noindent{\textbf{Limitation and future work} }   \ \
One limitation of our study is that we solely focus on regularization-based CIL algorithms. Given the recent advancements in model expansion-based methods and the prevalence of CIL studies utilizing pretrained models, it may be necessary to conduct similarly diverse evaluations in these areas as well. We leave this as a consideration for future work.

\section*{Acknowledgements}

This work was supported in part by the National Research Foundation of Korea (NRF) grant [No.2021R1A2C2007884] and by Institute of Information \& communications Technology Planning \& Evaluation (IITP) grants
[RS-2021-II211343, RS-2021-II212068, RS-2022-II220113,
RS-2022-II220959] funded by the Korean government (MSIT). It was also supported by SNU-NAVER Hyperscale AI Center and AOARD Grant No. FA2386-23-1-4079. 
\clearpage

\appendix
\pdfoutput=1

\section{Detailed Experimental Settings}
\noindent\textbf{Experimental settings of CIL algorithm} \ \
We achieve the result of CIL algorithms, FT, MAS~\citep{(MAS)aljundi2018memory} and LWF~\citep{(lwf)li2017learning} by implementing the CIL framework code proposed by \citep{(cil_survey)masana2020class}. We do not modify the default hyperparameters for each algorithm. We train these algorithms for 100 epochs for each task using the SGD optimizer with an initial learning rate of 0.1, momentum of 0.9, and weight decay of 0.0001. We also set a learning rate schedule that dropped the learning rate by a factor of 0.1 at 40 and 80 epochs, respectively. For all experiments, we use a mini-batch size of 256. We employed random sampling as the sampling algorithm for the exemplar memory.


We evaluate several regularization state-of-the-art CIL algorithms, including PODNet~\citep{(podnet)douillard2020podnet}, SSIL~\citep{(ss-il)ahn2021ss}, and AFC~\citep{(afc)kang2022class}. To ensure fair comparisons, we run the official code for each algorithm without modifying not only the default hyperparameters but also other settings for training, such as learning rate, epochs, and mini-batch size. Furthermore, we obtain experimental results for LUCIR~\citep{(lucir)hou2019learning} and BiC~\citep{(bic)wu2019large} using the code implemented in \citep{(podnet)douillard2020podnet}, also without any modification.

\noindent\textbf{Linear probing} \ \
We retrain the output layer while freezing the encoder. Specifically, we train the output layer for 30 epochs using a mini-batch size of 256, and utilize the SGD optimizer with an initial learning rate of 0.1, momentum of 0.9, and weight decay of 0.0001. We implement a learning rate schedule, which decreases the learning rate by a factor of 0.1 at the 10th and 20th epochs, respectively.

\noindent\textbf{$k$-NN evaluation} \ \
For all experiments, we utilize the $k$-NN implementation ($k=20$) provided by scikit-learn~\cite{pedregosa2011scikit}. In the classification process, we first fit the $k$-NN with the outputs of the encoder for the given inputs, and subsequently classify the test data using the $k$-NN classifier.

\noindent\textbf{Three downstream tasks} \ \
We select CIFAR-10~\citep{(cifar)krizhevsky2009learning}, STL-10~\citep{(stl10)coates2011analysis}, and CUB-200~\citep{(cub200)wah2011caltech} as downstream tasks for out-of-domain evaluation. For CIFAR-10, we randomly select 5,000 training images from the entire training dataset and resized the input images to $96\times96$ pixels. For STL-10 and CUB-200, we use the entire training dataset and maintained their original image sizes. We train only a newly added output layer while freezing the encoder. For CIFAR-10 and CUB-200, we train the output layer for 100 epochs using a mini-batch size of 128 and use SGD optimizer with an initial learning rate of 0.1, a momentum of 0.9, and a weight decay of 0.0001. We set the learning rate schedule to drop the learning rate by a factor of 0.1 at 40 and 80 epochs, respectively. In the case of STL-10, we change the number of epochs to 10 and the initial learning rate to 0.005.

\noindent\textbf{Experimental settings for unifying the base task's model} \ \
To unify the representation quality of the first task, we only reduce the number of epochs for the first task training of CIL algorithms that learn relatively better representations than others. The used number of training epochs for the first task is shown in Table \ref{table:reduced_epochs}.

\begin{table}[h]
\caption{The used numnber of training epochs for 10- and 11-tasks scenarios.}
\label{table:reduced_epochs}
\centering
\smallskip\noindent
\vspace{-.2in}
\resizebox{0.4\linewidth}{!}{
\begin{tabular}{|c||cc|cc|}
\hline
                                                            & \multicolumn{2}{c|}{10-tasks}   & \multicolumn{2}{c|}{11-tasks}     \\ \hline
                                                            & \multicolumn{1}{c|}{SSIL} & AFC & \multicolumn{1}{c|}{PODNet} & AFC \\ \hline \hline
\begin{tabular}[c]{@{}c@{}}First task\\ Epochs\end{tabular} & \multicolumn{1}{c|}{50}   & 45  & \multicolumn{1}{c|}{60}     & 45  \\ \hline
\end{tabular}
}
\end{table}

\section{Additional Experimental Results}

\subsection{Experimental analysis for other CIL algorithms}

\begin{figure*}[h]
\centering 
\subfigure[10-tasks, {Acc}($t$).]
{\includegraphics[width=0.32\linewidth]{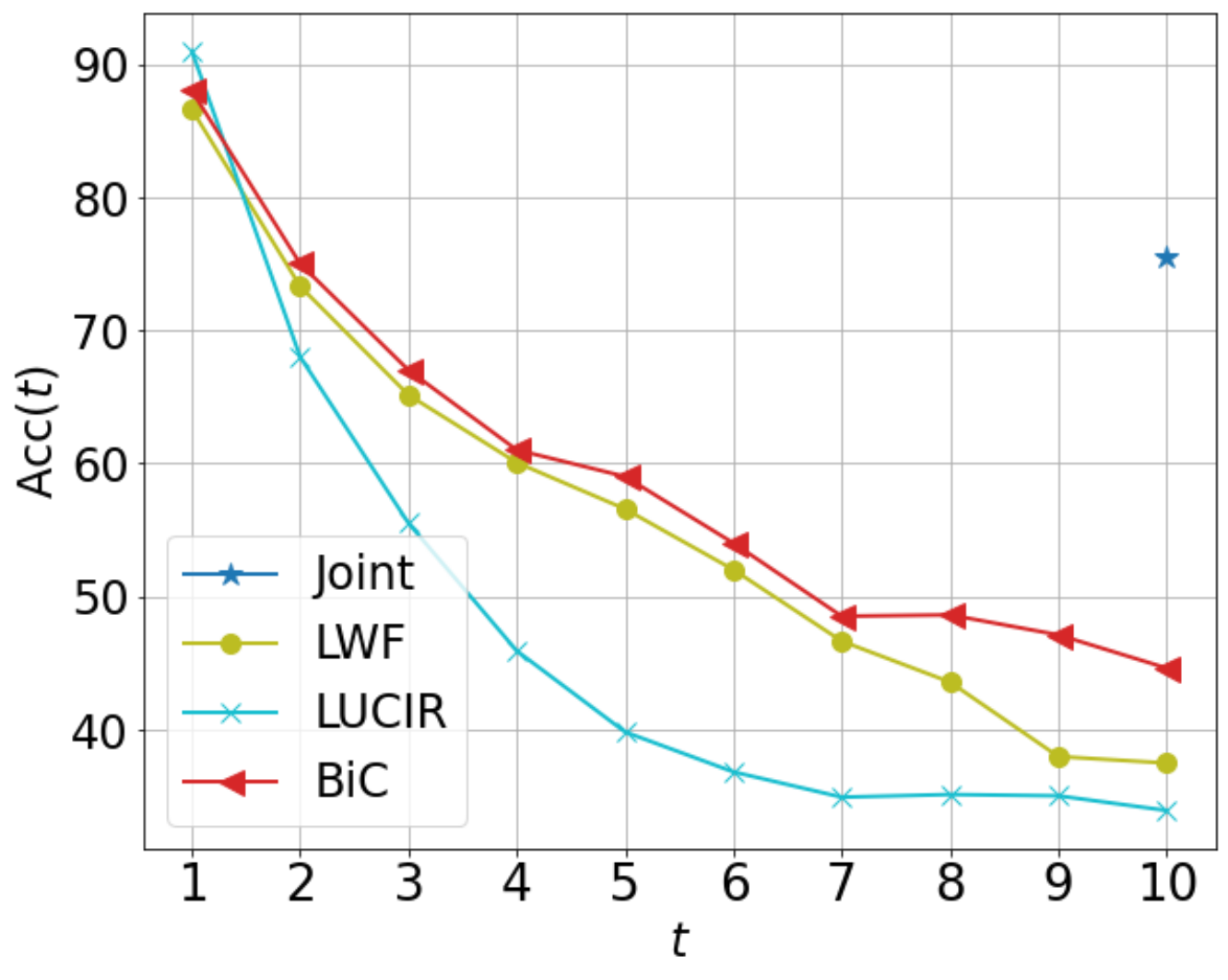}
\label{figure:10tasks_acc_additional}}
\subfigure[10-tasks, $k$\textbf{-NN}($t$)]
{\includegraphics[width=0.32 \linewidth]{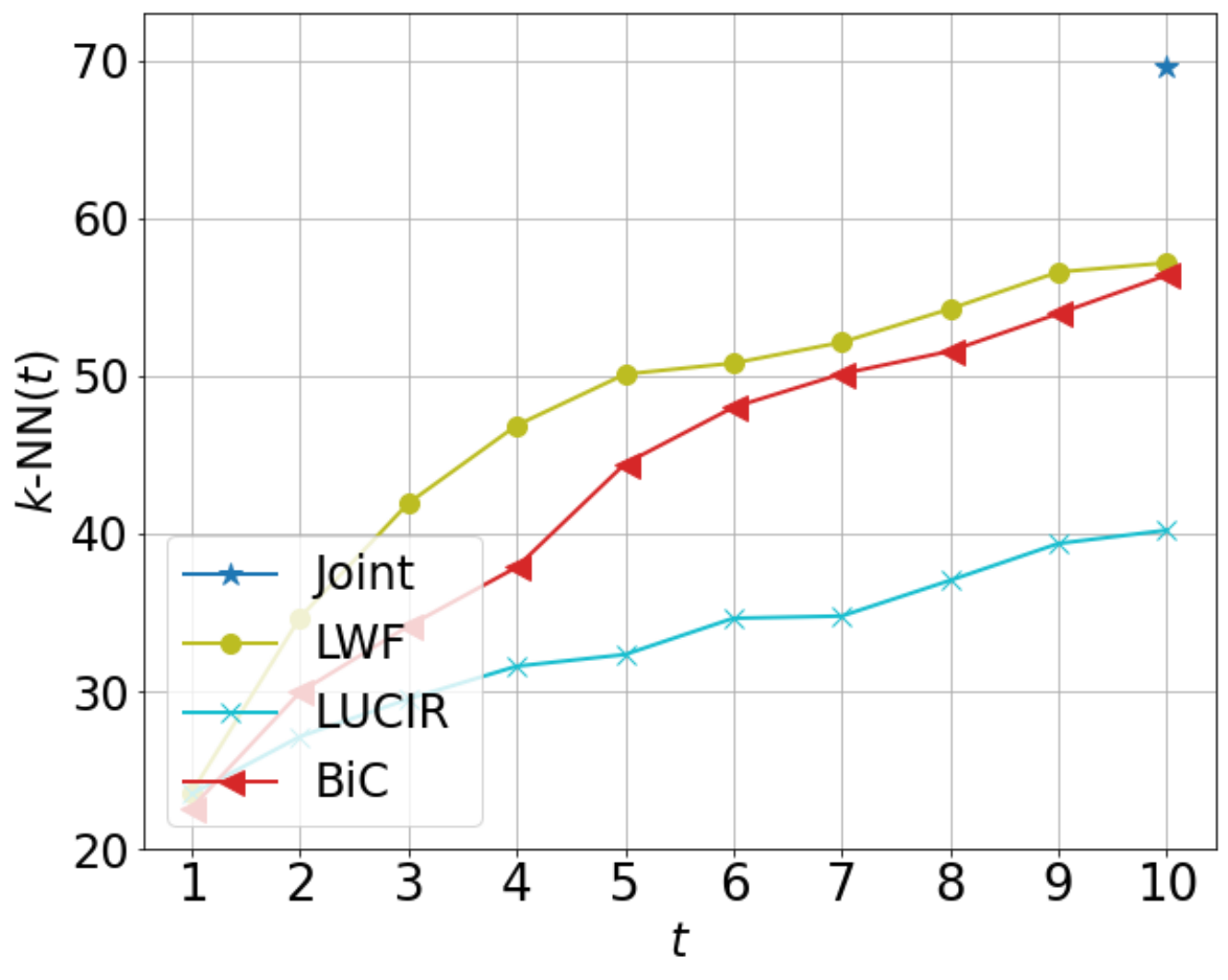}
\label{figure:10tasks_knn_additional}}
\subfigure[10-tasks, All eval. metrics]
{\includegraphics[width=0.32 \linewidth]{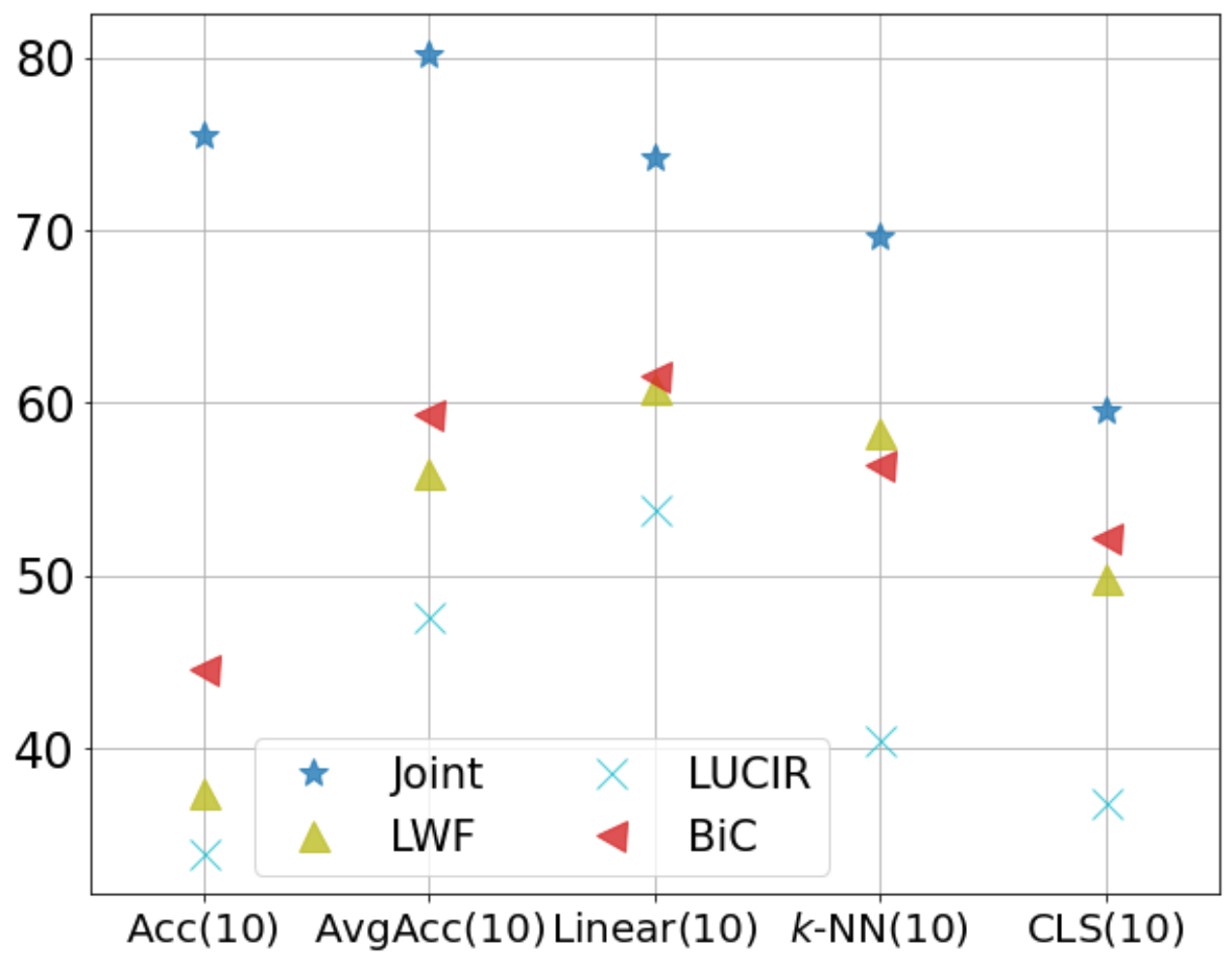}
\label{figure:10tasks_entire_additional}}
\subfigure[11-tasks, {Acc}($t$).]
{\includegraphics[width=0.32\linewidth]{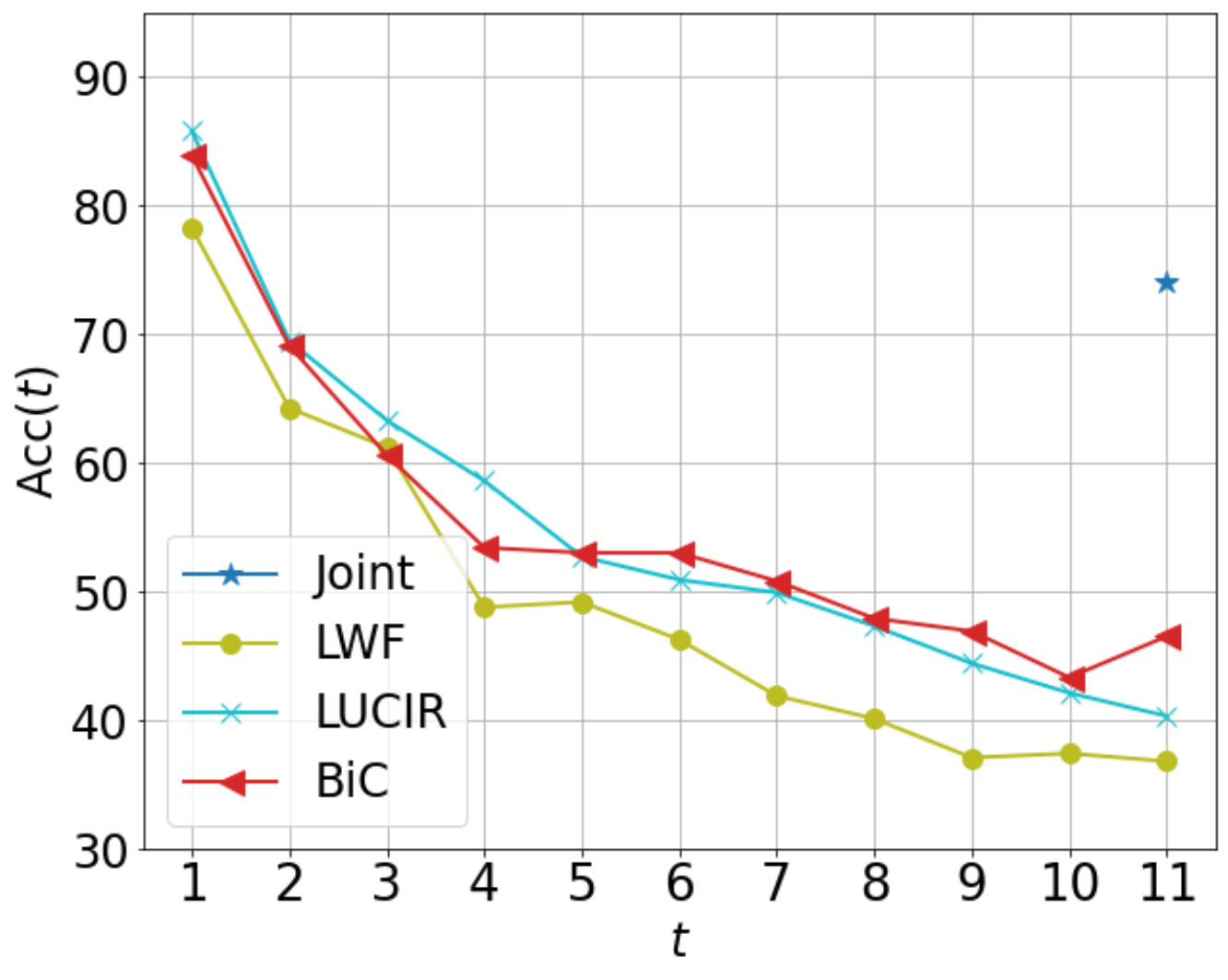}
\label{figure:11tasks_acc_additional}}
\subfigure[11-tasks, $k$\textbf{-NN}($t$)]
{\includegraphics[width=0.32 \linewidth]{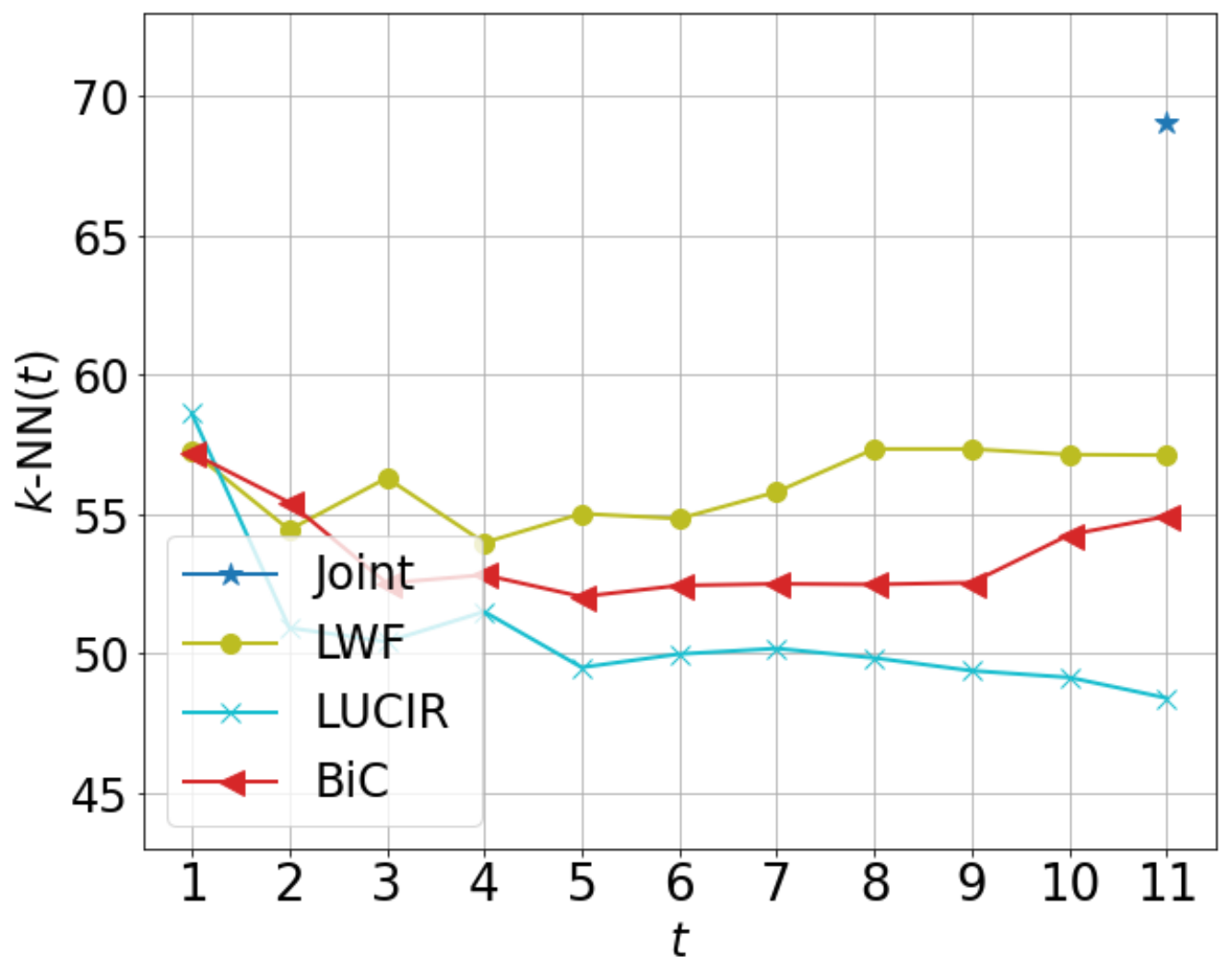}
\label{figure:11tasks_knn_additional}}
\subfigure[11-tasks, All eval. metrics]
{\includegraphics[width=0.32 \linewidth]{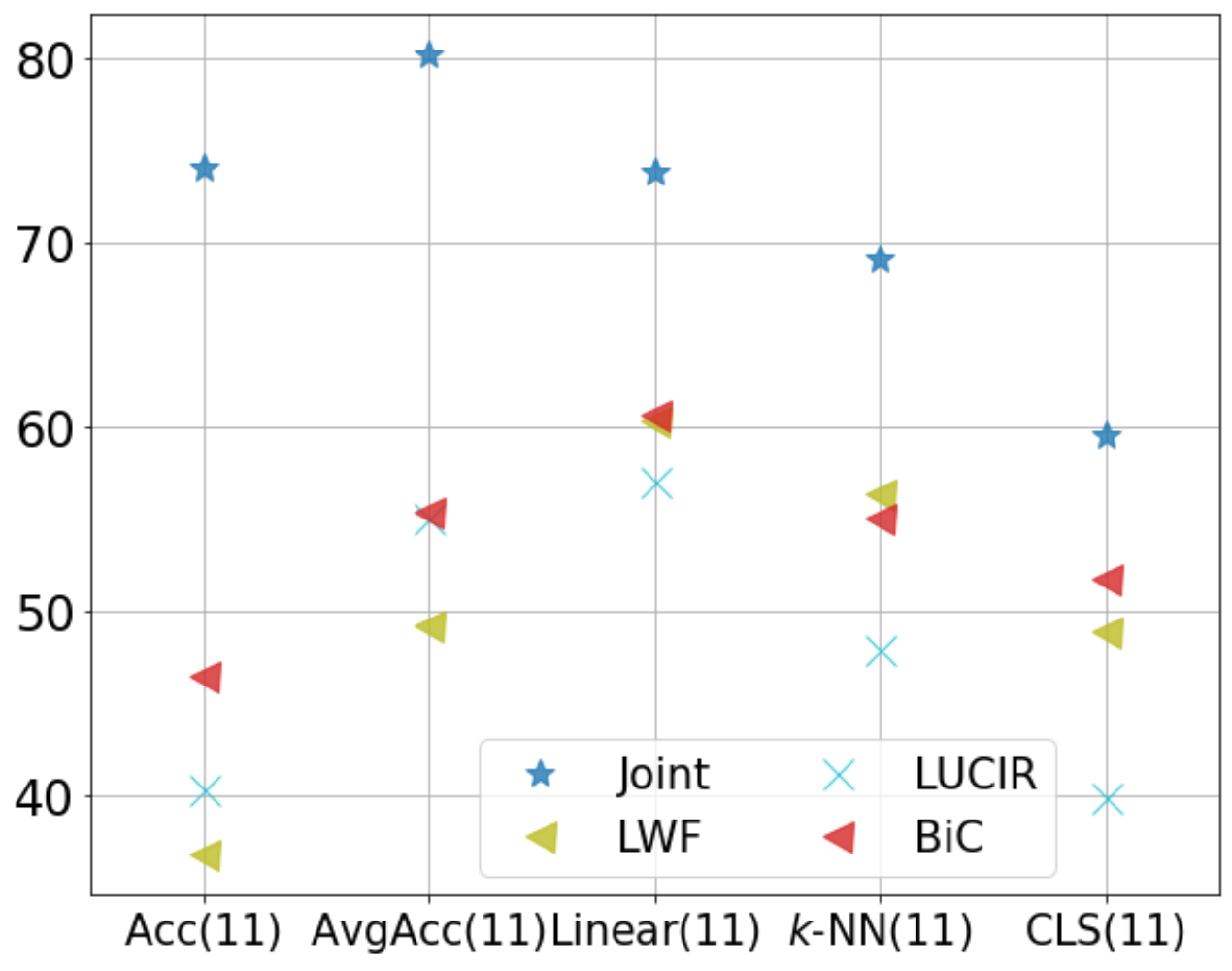}
\label{figure:11tasks_entire_additional}}
\vspace{-.1in}
\caption{Additional experimental results of LWF and LUCIR for the scenario of 10-tasks and 11-tasks.}
\label{figure:additional_10_11_tasks}
\end{figure*}

In our experiment using the ImageNet-100 dataset, we conduct additional experiments on LWF~\citep{(lwf)li2017learning} and LUCIR~\citep{(lucir)hou2019learning}, and the results are shown in Figure \ref{figure:additional_10_11_tasks}. The results for BiC~\citep{(bic)wu2019large} are added for comparison and are consistent with the results in the manuscript. We are able to confirm experimental results similar to the findings in the manuscript in two scenarios (10-tasks and 11-tasks). 
First, since the learning of the BiC encoder is carried out through a form of knowledge distillation similar to LWF, the evaluation results for representation quality between LWF and BiC show no significant difference. 
Second, in the case of LUCIR, although it achieved higher \textbf{Acc}($t$) and \textbf{AvgAcc}($t$) than LWF in the situation of learning from the base model (11-tasks), it demonstrates that the representation quality learned by this algorithm is significantly inferior to LWF.

\begin{figure*}[!h]
\centering 
\subfigure[FT (10-tasks)]
{\includegraphics[width=0.32\linewidth]{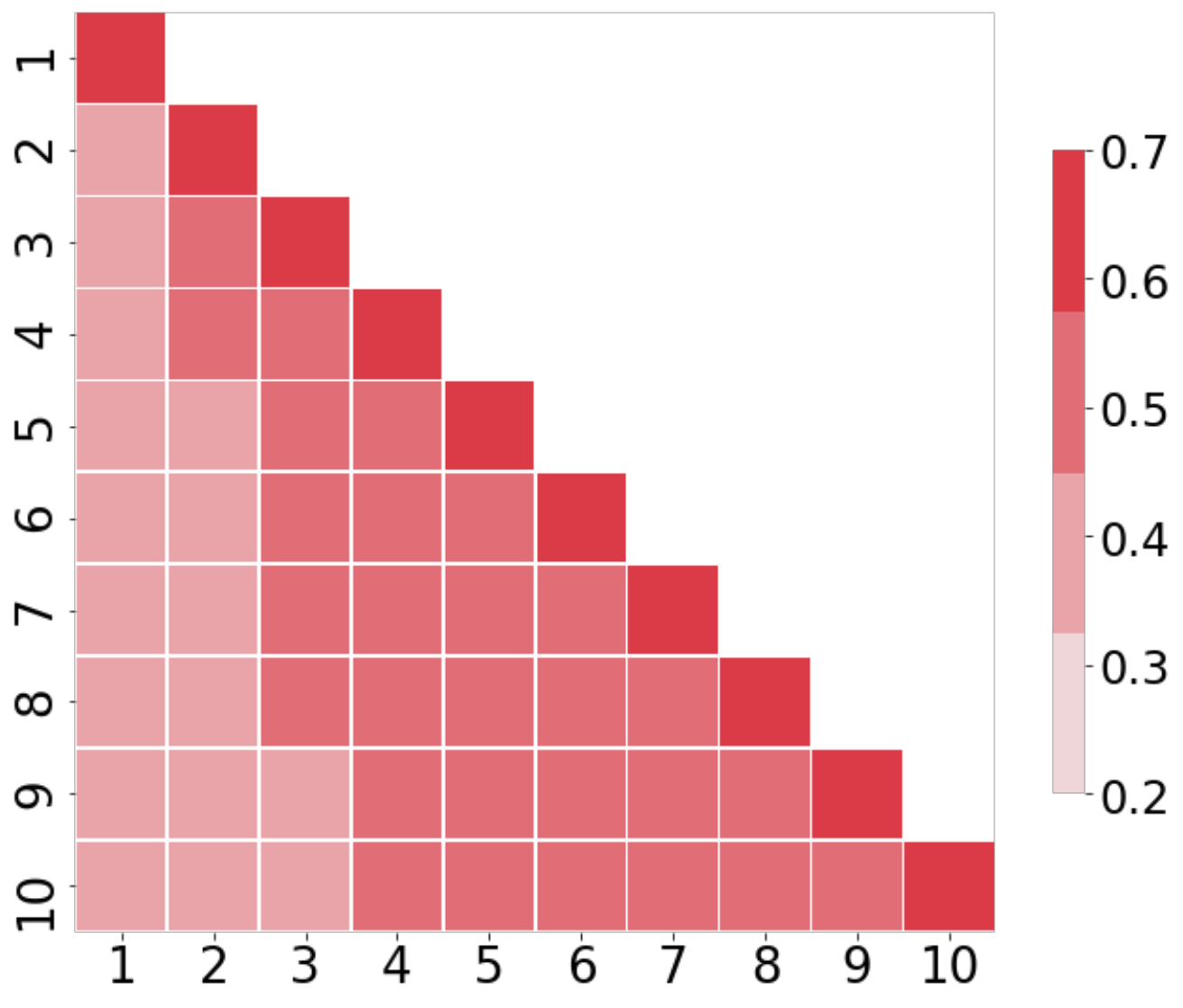}
\label{figure:10tasks_cka_ft}}
\subfigure[MAS (10-tasks)]
{\includegraphics[width=0.32 \linewidth]{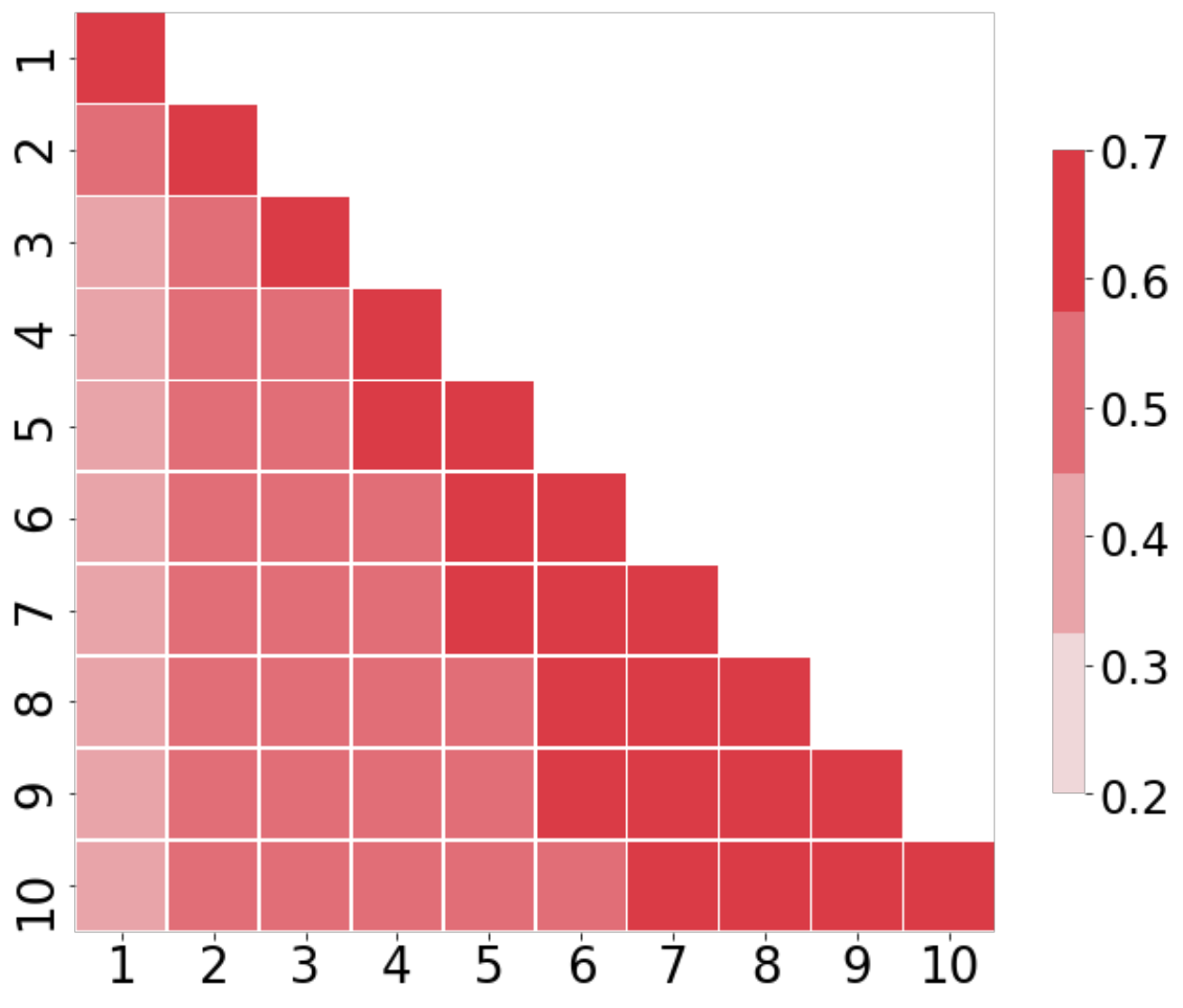}
\label{figure:10tasks_cka_mas}}
\subfigure[SSIL (10-tasks)]
{\includegraphics[width=0.32\linewidth]{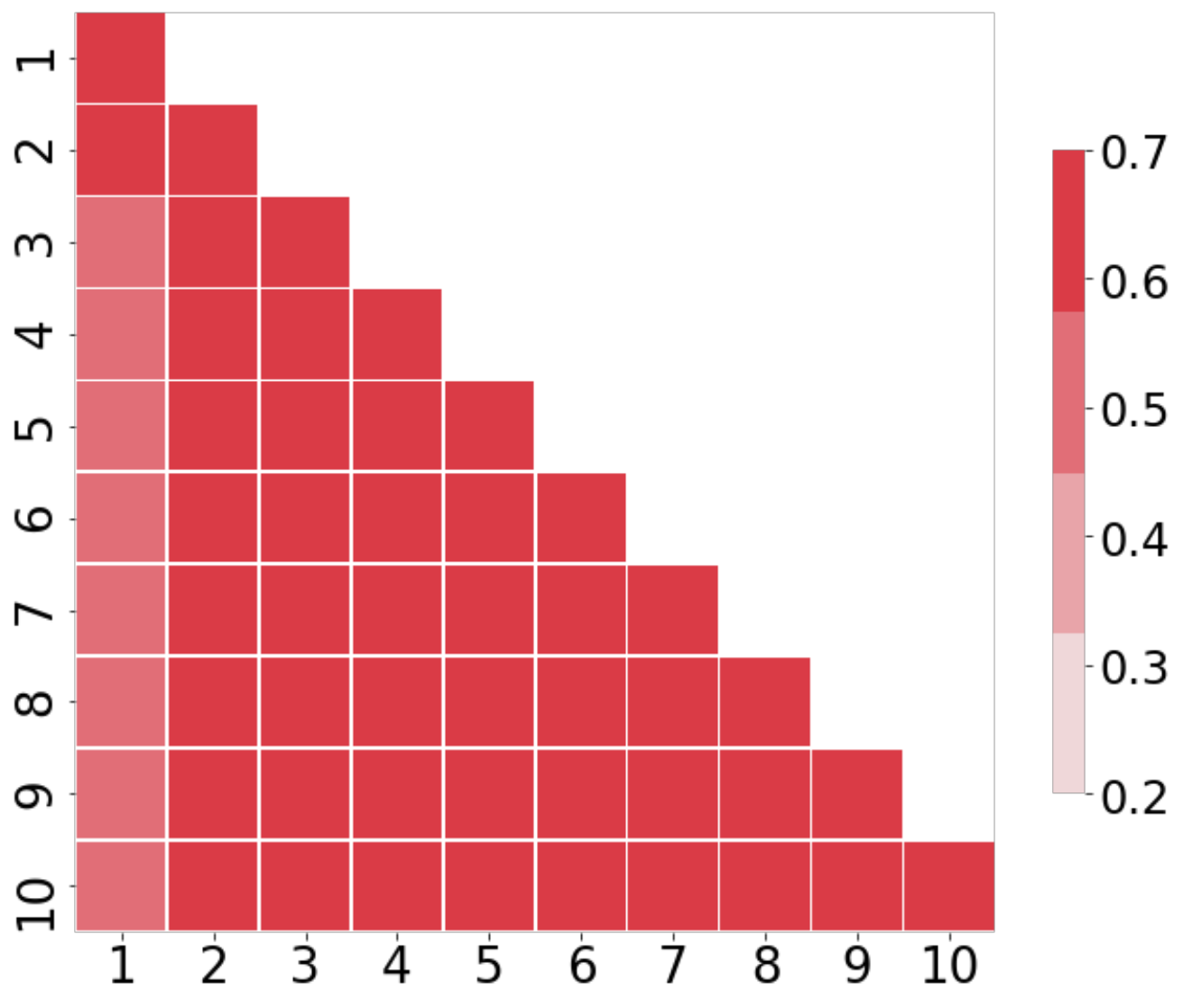}
\label{figure:10tasks_cka_ssil}}
\subfigure[FT (11-tasks)]
{\includegraphics[width=0.32\linewidth]{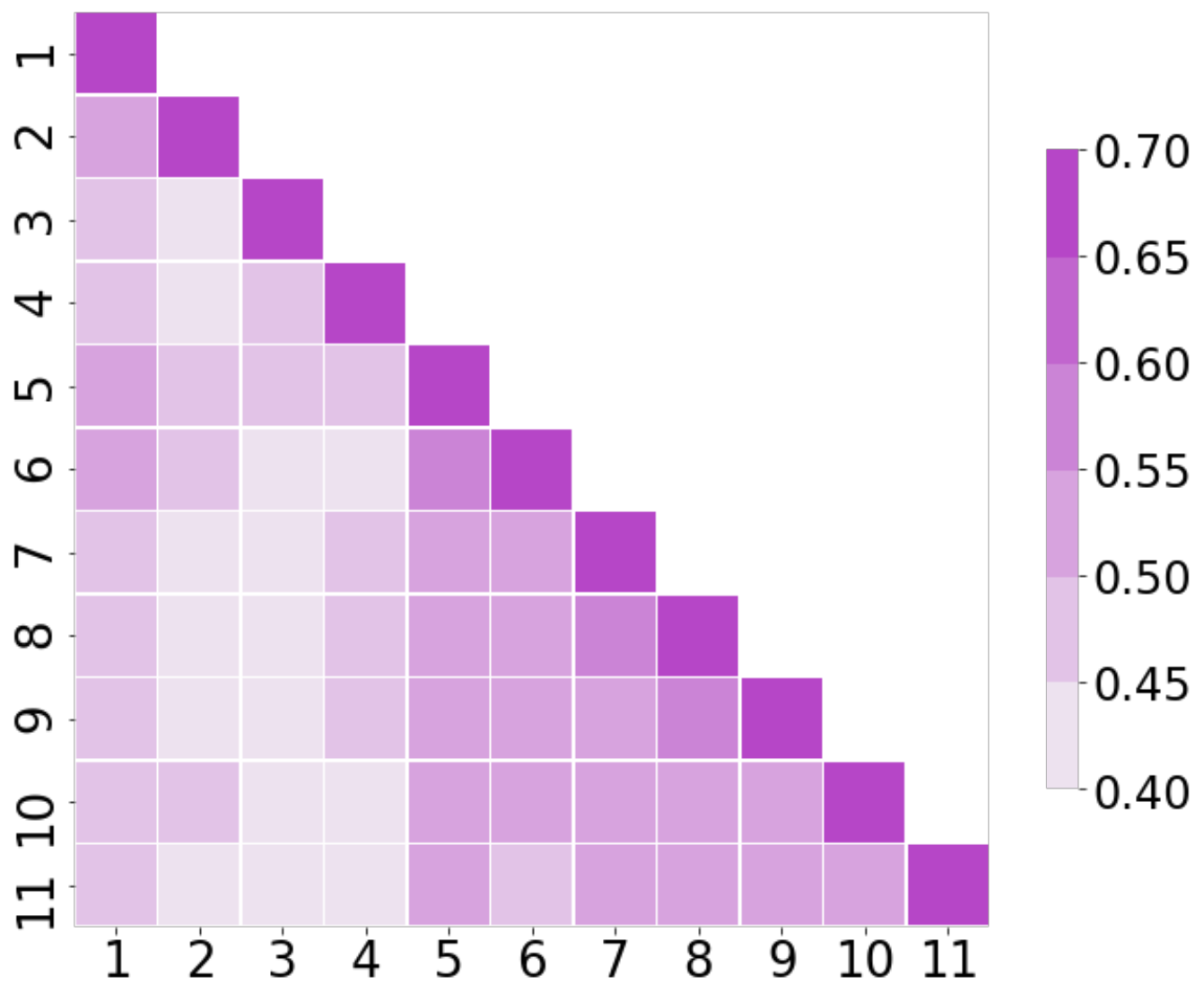}
\label{figure:11tasks_cka_ft}}
\subfigure[MAS (11-tasks)]
{\includegraphics[width=0.32 \linewidth]{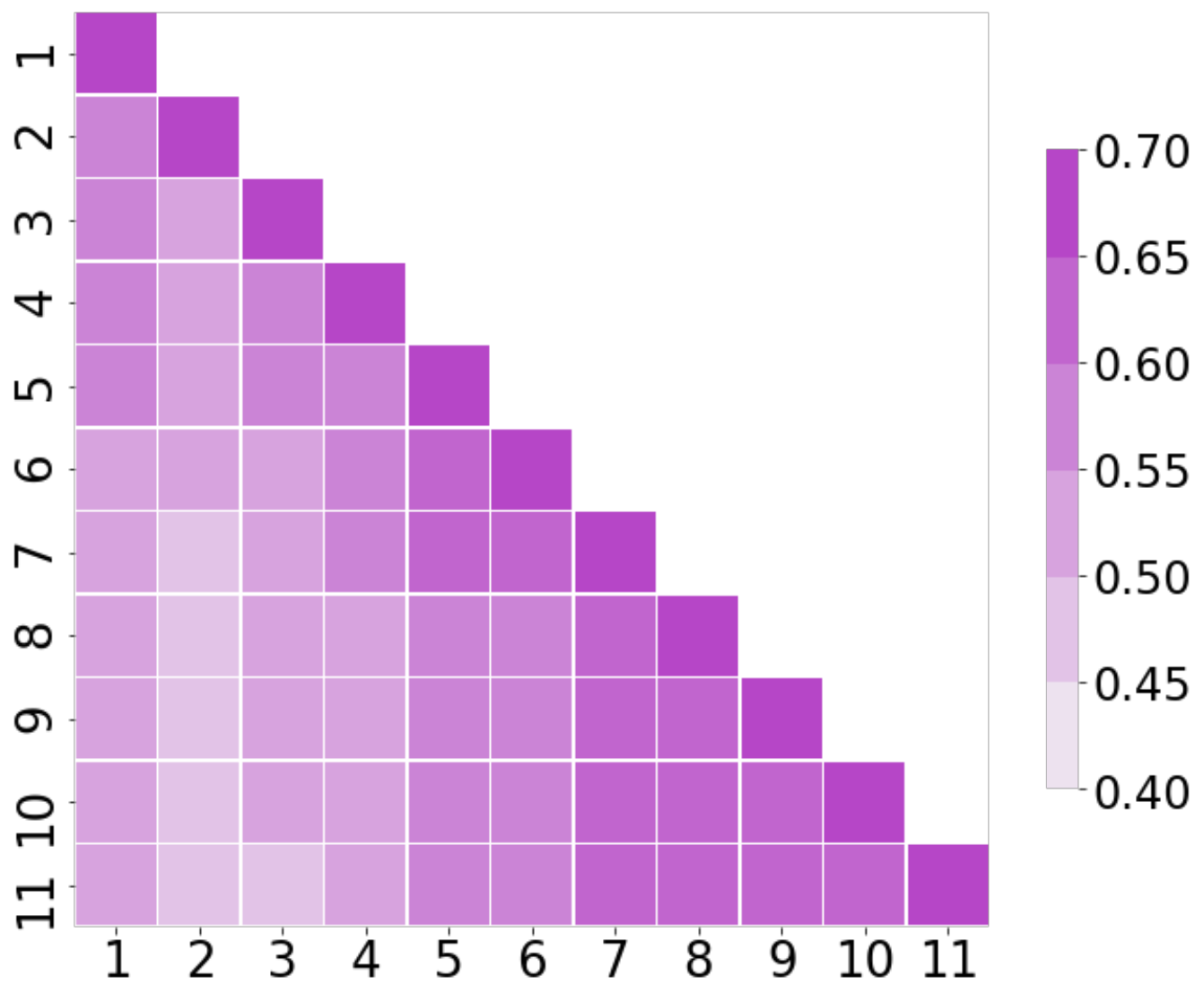}
\label{figure:11tasks_cka_mas}}
\subfigure[SSIL (11-tasks)]
{\includegraphics[width=0.32 \linewidth]{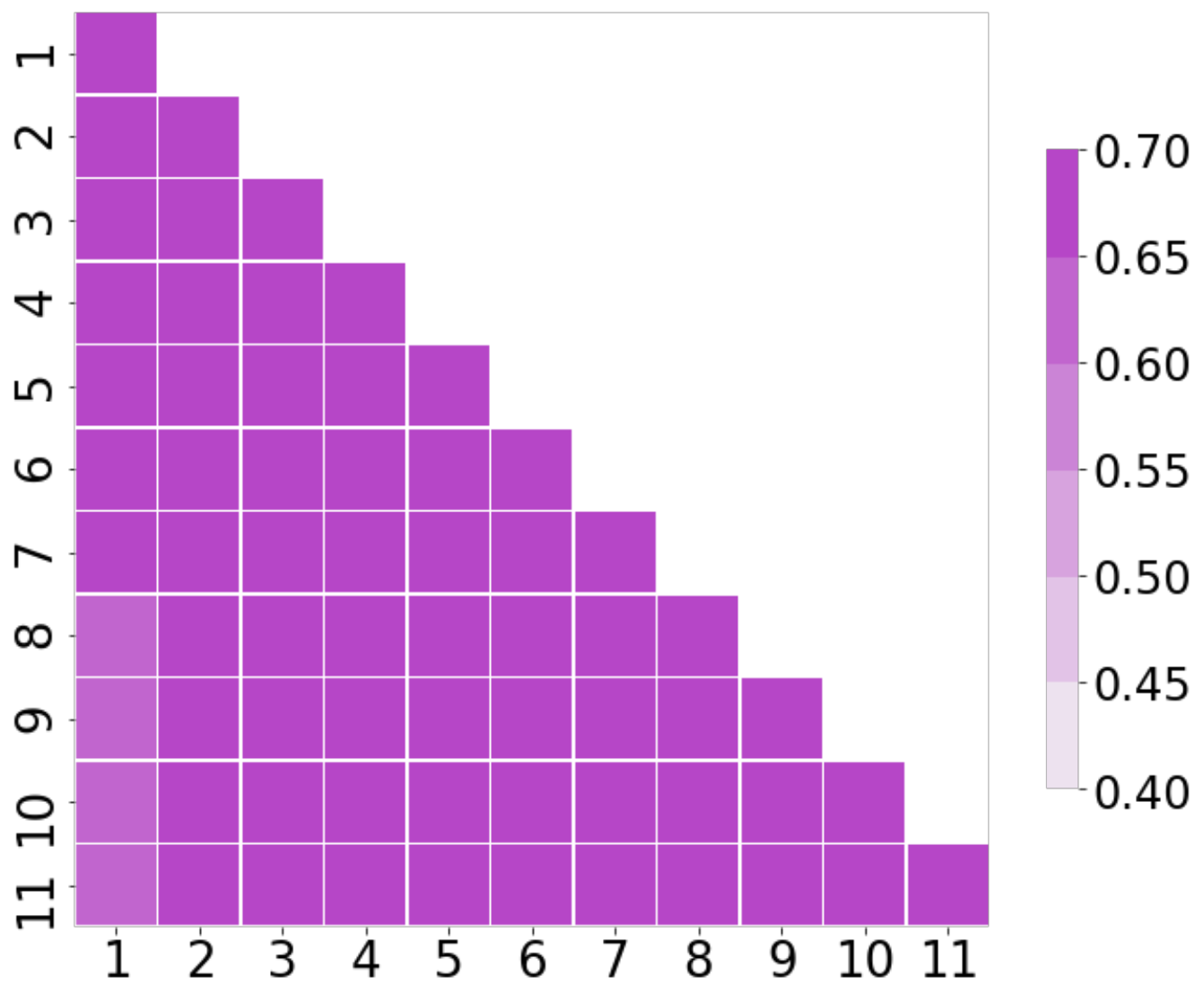}
\label{figure:11tasks_cka_ssil}}
\vspace{-.1in}
\caption{{CKA}${(t_{1},t_{2})}$ in 10-tasks and 11-tasks scenario.}
\vspace{-.1in}
\label{figure:additional_10tasks_cka}
\end{figure*}

\subsection{CKA results for other CIL algorithms}

Figure \ref{figure:additional_10tasks_cka} shows the {\textbf{CKA}}${(t_{1},t_{2})}$ results for other algorithms in the experiment using ImageNet-100. In the case of 10-tasks, first, FT and MAS show results similar to the Joint in the manuscript. Second, SSIL maintains a relatively strong similarity. 
Through this, we can further confirm that each SOTA algorithm focuses on stability to prevent significant changes in learned representations, leading to poorer representation learning compared to FT and MAS in the 10-tasks scenario. 
In the case of 11-tasks, since a relatively large amount of knowledge (half of the total class number) is learned in the first task, algorithms that focus on stability can achieve relatively superior results. 
Taking this into account, FT and MAS show significant changes in representation, while SSIL does not, showing a similar trend to the evaluation results for representation quality for each algorithm (See Figure \ref{figure:11tasks_figures}).

\begin{figure*}[h]
\centering 
\subfigure[{Acc}($t$).]
{\includegraphics[width=0.32\linewidth]{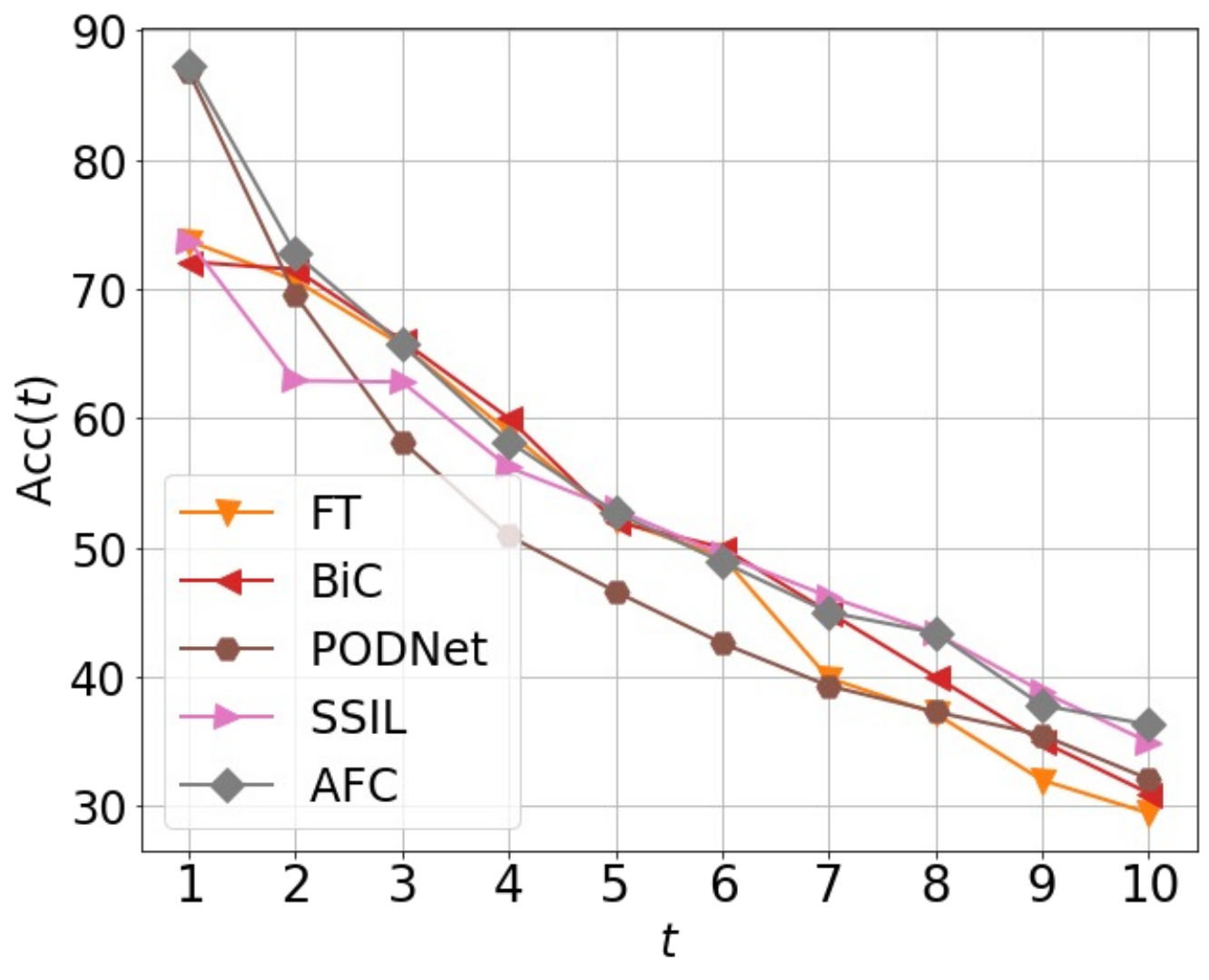}
\label{figure:cifar100_10tasks_acc}}
\subfigure[$k$\textbf{-NN}($t$)]
{\includegraphics[width=0.32 \linewidth]{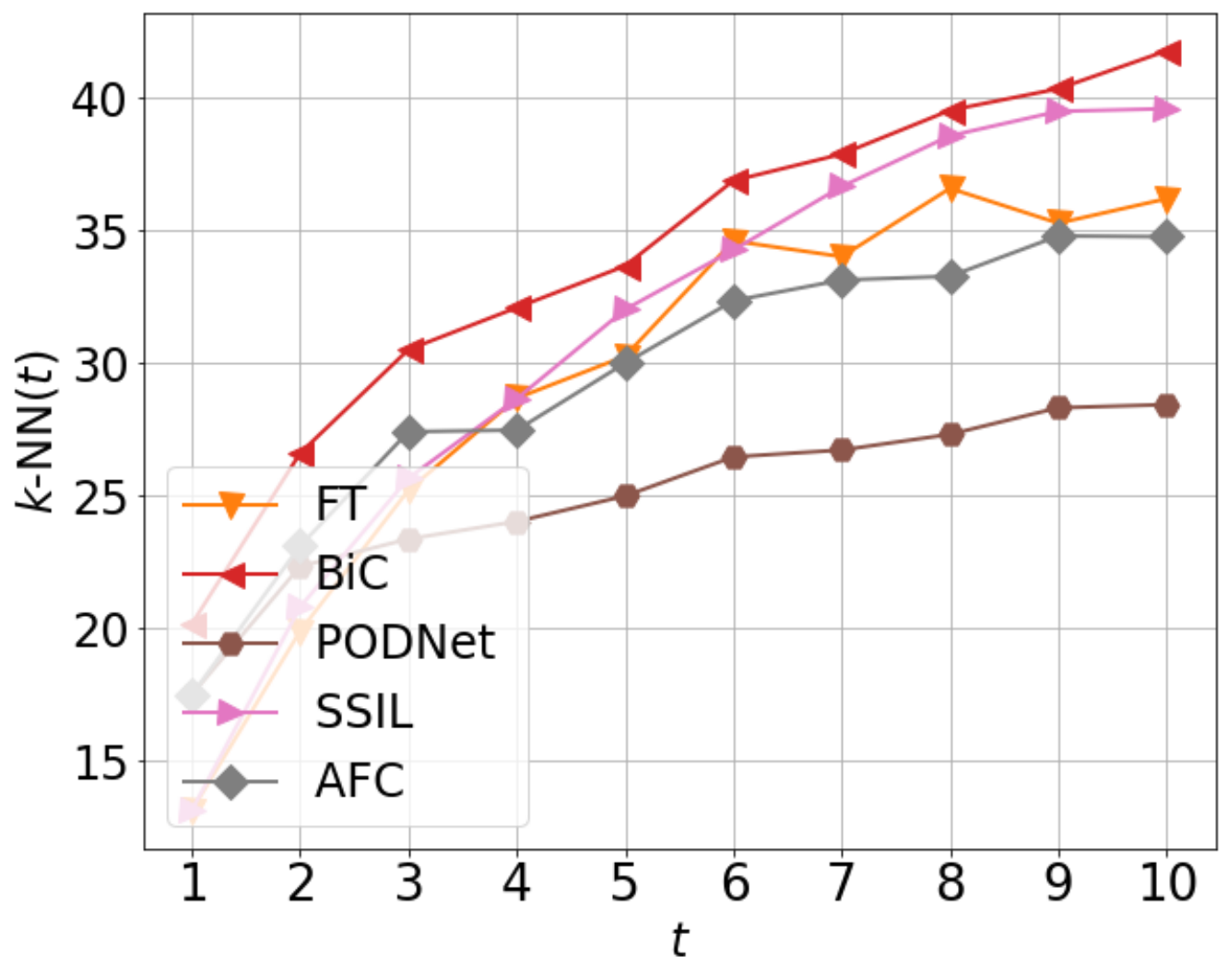}
\label{figure:cifar100_10tasks_knn}}
\subfigure[All eval. metrics]
{\includegraphics[width=0.32 \linewidth]{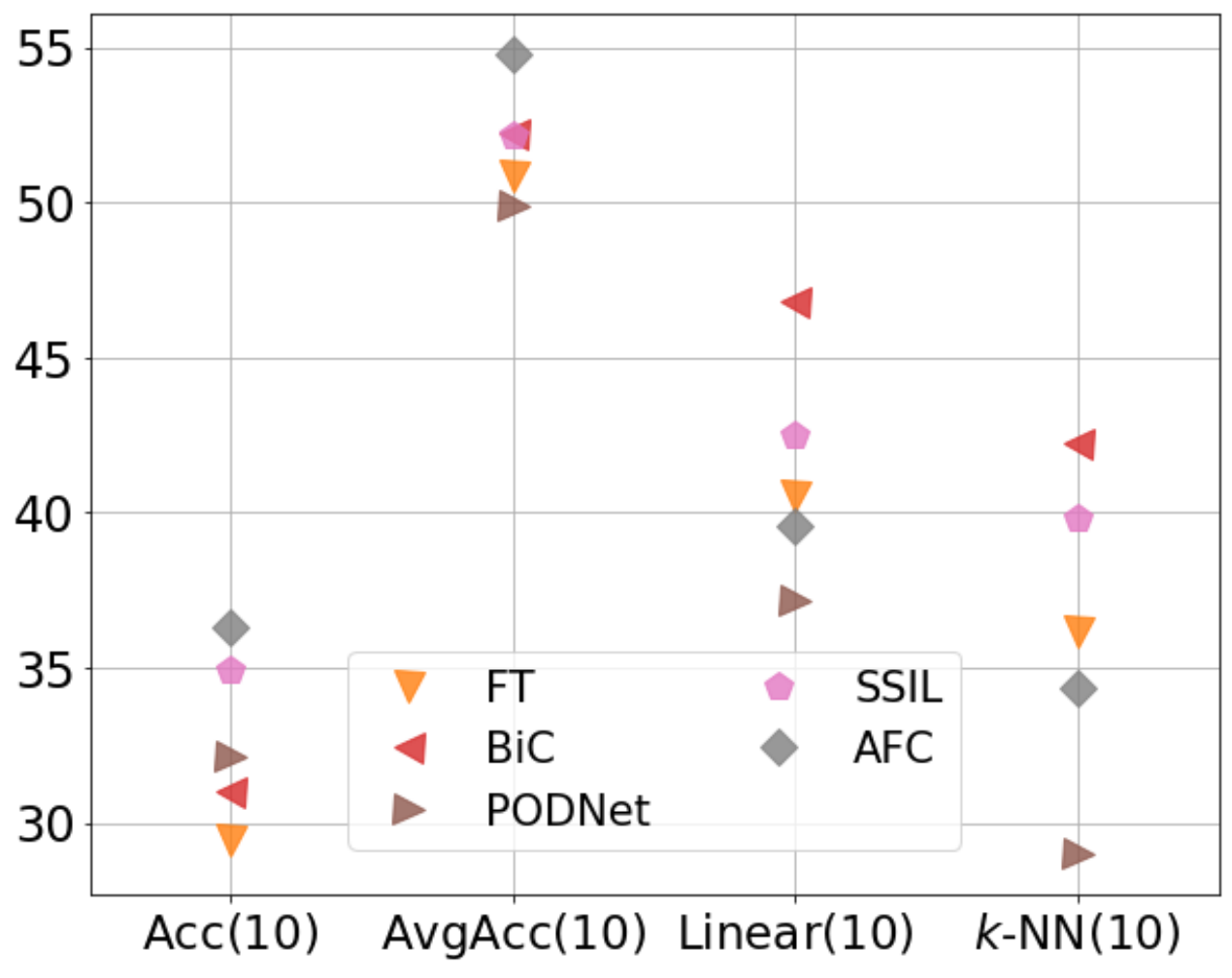}
\label{figure:cifar100_10tasks_entire}}
\vspace{-.1in}
\caption{Additional experimental results with CIFAR-100 for the scenario of 10-tasks.}
\vspace{-.1in}
\label{figure:additional_cifar100}
\end{figure*}

\subsection{Experimental analysis with CIFAR-100}

To investigate whether the analysis results proposed in the manuscript are dataset-dependent, we conduct CIL experiments using CIFAR-100  for the 10-tasks scenario. We use five representative algorithms (FT, BiC, PODNet, SSIL, and AFC) and conduct experiments with their reported default hyperparameters. 
All experiments are conducted using the ResNet-18 model, and for SSIL, which has no experiments on CIFAR-100 in the their paper, we apply the same number of epochs (160) used for training in PODNet and AFC. 
Additionally, we only conduct in-domain evaluation.

Figure \ref{figure:additional_cifar100} shows the experimental results, and we obtain analysis results similar to those of ImageNet-100 in the manuscript. 
First, we confirm that AFC and SSIL achiev relatively superior \textbf{Acc}($t$) and \textbf{AvgAcc}($t$), but learned representations are inferior to those learned by BiC. Second, we observe that PODNet learns significantly worse representations compared to other algorithms on CIFAR-100.


\newpage






\bibliography{collas2024_conference}
\bibliographystyle{collas2024_conference}

\end{document}